\documentclass[12pt,logo]{spell}
\usepackage{nicefrac}           

\usepackage{multirow}
\usepackage{array}
\usepackage{adjustbox}
\usepackage{makecell}
\usepackage{longtable}
\usepackage{arydshln}           

\usepackage{wrapfig}
\usepackage{subcaption}
\usepackage{rotating}
\usepackage[abs]{overpic}
\usepackage{tikz}               

\usepackage[most]{tcolorbox}

\usepackage{listings}
\usepackage{csquotes}

\usepackage{fontawesome}

\usepackage[capitalize]{cleveref}

\usepackage{tocloft}

\usepackage{lipsum}

\usepackage{doi}

\renewcommand{\phi}{\varphi}

\renewcommand{\geq}{\geqslant}

\renewcommand{\epsilon}{\varepsilon}
\renewcommand{\imath}{\mathrm{i}}

\newlength{\restsubwidth}
\newlength{\restsubheight}
\newlength{\restsubmoreheight}
\newcommand{\rest}[2]{%
        \settowidth{\restsubwidth}{\ensuremath{#2}}
        \settoheight{\restsubheight}{\ensuremath{{}_{#2}}}
        \ensuremath{{#1\hskip 0.5pt}_{\vrule\kern2pt\parbox[b][%
        4pt][b]{\the\restsubwidth}{%
                        \ensuremath{{}_{#2}}}}}
        }

\def\dataset{{\textit{EgoGroups}}}

\NewDocumentCommand{\paola}{ mO{} }{\noindent\textcolor{purple}{\textsf{\small[#1]}\textsuperscript{\textit{Paola}}}}

\NewDocumentCommand{\jeffri}{ mO{} }{\noindent\textcolor{brown}{\textsf{\small[#1]}\textsuperscript{\textit{Jeffri}}}}

\NewDocumentCommand{\guha}{ mO{} }{\noindent\textcolor{red}{\textsf{\small[#1]}\textsuperscript{\textit{Guha}}}}

\NewDocumentCommand{\pranav}{ mO{} }{\noindent\textcolor{violet}{\textsf{\small[#1]}\textsuperscript{\textit{Pranav}}}}

\newcommand{\red}[1]{\textcolor{red}{#1}}
\newcommand{\blue}[1]{\textcolor{blue}{#1}}

\definecolor{bluelink}{RGB}{0,113,188}
\definecolor{greenlink}{RGB}{0,188,113}
\definecolor{linkteal}{RGB}{0,102,102}
\definecolor{linkblue}{RGB}{0,51,102}
\hypersetup{
    colorlinks=true,
    citecolor=green!93!black,
    linkcolor=linkblue,
    urlcolor=bluelink
}

\definecolor{codekeyword}{rgb}{0.0, 0.0, 0.5}
\definecolor{codecomment}{rgb}{0.0, 0.5, 0.0}
\definecolor{codestring}{rgb}{0.56, 0.0, 1.0}

\lstdefinestyle{pythonstyle}{
    language=Python,
    basicstyle=\ttfamily\small,
    keywordstyle=\color{codekeyword}\bfseries,
    commentstyle=\color{codecomment}\itshape,
    stringstyle=\color{codestring},
    showstringspaces=false,
    breaklines=true,
    tabsize=4,
    numbers=none,
    frame=none,
    backgroundcolor=\color{white},
    captionpos=b,
    morekeywords={self, __init__, __name__, __main__},
}
\title{\center EgoGroups: A Benchmark For Detecting \\Social Groups of People in the Wild}

\author{Jeffri Murrugarra-Llerena$^{1}$, Pranav Chitale$^{1}$, Zicheng Liu$^{1}$, Kai Ao$^{1}$, Yujin Ham$^{2}$,\\
\vspace{-12pt}
\normalfont{
Guha Balakrishnan$^{2}$, Paola Cascante-Bonilla$^{1}$}\\
    \vspace{8pt}
    $^{1}$Stony Brook University ~~~ $^{2}$Rice University\\
    \vspace{8pt}
    {\small
    \faGlobe\ \href{https://lab-spell.github.io/EgoGroups/}{Project Page} \quad
    \faGithub\ \href{https://github.com/lab-spell/EgoGroups}{Repository}
    }
}

\begin{abstract}
Social group detection, or the identification of humans involved in reciprocal interpersonal interactions (e.g., family members, friends, and customers and merchants), is a crucial component of social intelligence needed for agents transacting in the world. The few existing benchmarks for social group detection are limited by low scene diversity and reliance on third-person camera sources (e.g., surveillance footage). Consequently, these benchmarks generally lack real-world evaluation on how groups form and evolve in diverse cultural contexts and unconstrained settings. To address this gap, \textbf{we introduce \dataset{}}, a first-person view dataset that captures social dynamics in cities around the world. \dataset{} spans 65 countries covering low, medium, and high-crowd settings under four weather/time-of-day conditions. We include dense human annotations for person and social groups, along with rich geographic and scene metadata. Using this dataset, we performed an extensive evaluation of state-of-the-art VLM/LLMs and supervised models on their group detection capabilities. We found several interesting findings, including VLMs and LLMs can outperform supervised baselines in a zero-shot setting, while crowd density and cultural regions clearly influence model performance.
\end{abstract}

\begin{document}

\maketitle

\begin{abstract}
Social group detection, or the identification of humans involved in reciprocal interpersonal interactions (e.g., family members, friends, and customers and merchants), is a crucial component of social intelligence needed for agents transacting in the world. The few existing benchmarks for social group detection are limited by low scene diversity and reliance on third-person camera sources (e.g., surveillance footage). Consequently, these benchmarks generally lack real-world evaluation on how groups form and evolve in diverse cultural contexts and unconstrained settings. To address this gap, \textbf{we introduce \dataset{}}, a first-person view dataset that captures social dynamics in cities around the world. \dataset{} spans 64 countries covering low, medium, and high-crowd settings under four weather/time-of-day conditions. We include dense human annotations for person and social groups, along with rich geographic and scene metadata. Using this dataset, we performed an extensive evaluation of state-of-the-art VLM/LLMs and supervised models on their group detection capabilities. We found several interesting findings, including VLMs and LLMs can outperform supervised baselines in a zero-shot setting, while crowd density and cultural regions clearly influence model performance.

\end{abstract}    
\section{Introduction}

Humans in public environments often interact and move in social groups. A social group consists of two or more persons reciprocally engaged in an interaction or activity, rather than simply being co-located in a space. Families shopping together, people conversing, and a customer and merchant exchanging goods are valid groupings, whereas strangers waiting at a bus stop, commuters walking in the same direction, and a concert crowd as a whole are not. Understanding group membership and actions is therefore fundamental to how we as humans navigate and interact with others in realistic environments.
As research advances toward the deployment of embodied agents and other assistive devices in real-world settings, it is crucial for these technologies to reason about how humans approach, join, or avoid groups from a first-person perspective (e.g., from camera-mounted perspective), rather than from fixed, calibrated camera arrays. While research on first-person camera-view datasets has made significant progress with respect to social attention and conversational cues~\cite{ego4D, li2025sekai}, they do not directly evaluate group membership, continuity, or group-level activities.

\begin{figure}[t] 
    \centering
    \includegraphics[width=0.9\linewidth]{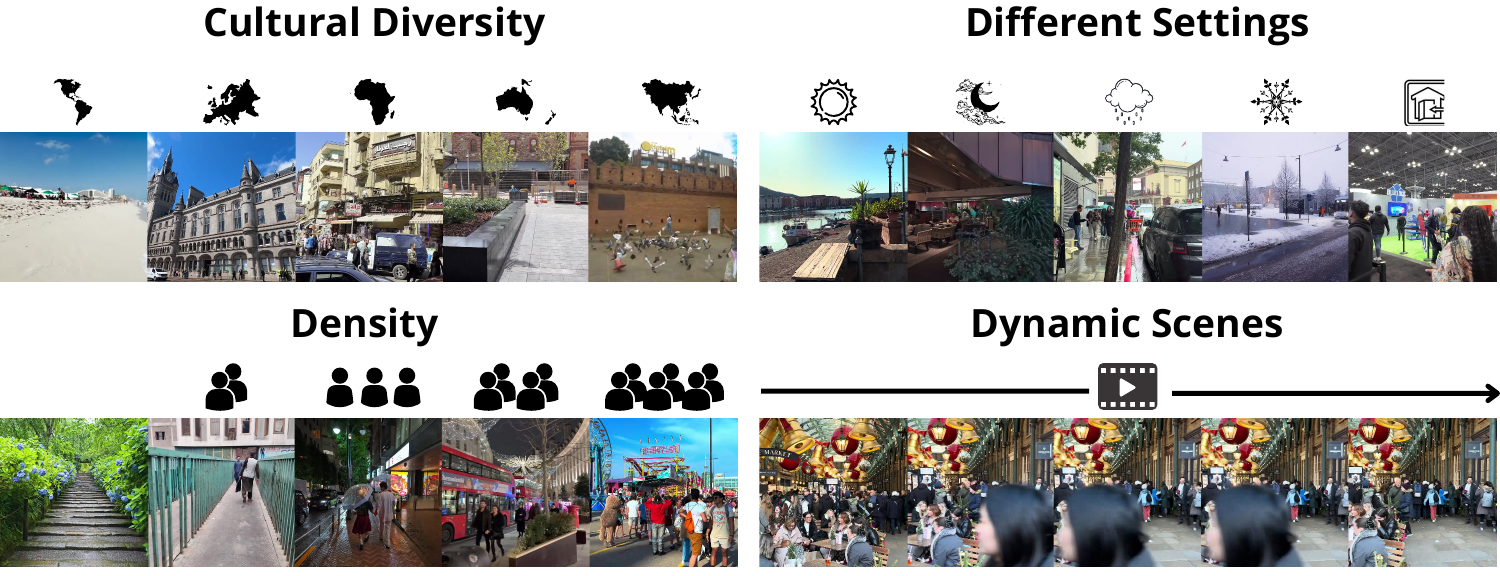}
    \caption{\textbf{\dataset{} is a dataset that captures social group behavior from walking-tour videos in cities around the world.} We show culturally diverse settings with large groups of people or isolated pedestrians, in videos with a  narrow, constantly moving field-of-view with frequent motion blur and occlusions. 
    }
    \label{fig:concept_figure}
\end{figure}

A core challenge in social group understanding is the inherent complexity of first-person visual perception. First-person videos exhibit rapid viewpoint change, motion blur, and frequent occlusions. Fields-of-view are often narrow and, when coupled with abrupt entry and exit of people in the camera's view, lead to dramatic variations in scale and focus based on proximity to the camera wearer. These factors degrade the positional and orientational cues (proximity, head/body pose) that underlie group reasoning and make frame-level classification unreliable without robust tracking and temporal aggregation. Moreover, scene context strongly influences the likelihood of group memberships. In particular, location and crowd density impact visual statistics and spacing norms, causing a failure in generalization of model performance across locations. Additionally, cultural context amplifies these shifts; interpersonal spacing and orientation norms vary measurably across countries and regions, altering what constitutes ``near,'' ``with'' or ``approaching'' in public. Prior works report systematic differences in preferred distances between individuals and groups of people around the globe depending on country and region~\cite{aiello1980personal,beaulieu2004intercultural,evans1973personal,lomranz1976cultural}. 

While recent years have recorded significant progress towards social group detection, existing benchmarks~\cite{salsa_dataset, PANDA, JRDB-Act, kim2024towards} 
primarily rely on third-person camera sources and are focused on specific scenarios. For instance, JRDB~\cite{JRDB-Act} annotates social groups observed from the perspective of a robot navigating a single location (Adelaide University campus), CAFE~\cite{kim2024towards} captures interactions exclusively in 6 cafes with four static cameras, and PANDA~\cite{PANDA} employs a static gigapixel camera in 21 distinct locations. 
Overall, these datasets are predominantly collected in a single city or country, reflecting a homogeneous demographic, consequently not accounting for the cultural variation in how social groups form and behave across different societies. Thus, our work is motivated by the lens of diversity and real-world representativeness. 

This study introduces \dataset{}, a dataset and benchmark centered on first-person walking tour videos in diverse urban settings and crowd densities around the world, designed to evaluate group detection under real-world camera motions and settings. \dataset{} spans over 65 countries, totaling 540 video clips (roughly $45$ minutes of total footage annotated from an initial raw video set of 16 hours) collected across a global range of cities and venue types. We took special care to select clips encompassing the full diversity of real-world scenarios, interactions, and group activities encountered in everyday life. For each video clip, we also provide fine-grained person detections, social group annotations, and geography/scene metadata (city, venue, density, and time-of-day), as illustrated in \autoref{fig:concept_figure} and~\ref{fig:annotation_examples}. Captured scenes range from nearly empty sidewalks to highly congested markets and transit hubs, with some frames containing up to 26 annotated social groups. By combining first-person view videos with variation in crowd density in various locales, \dataset{} reveals a variety of group configurations that are largely absent from existing first-person view and panoramic video benchmarks for group detection. Beyond group detection annotations, we also augment \dataset{} with computed statistics of crowd density, time-of-day, indoor/outdoor context, weather, and cultural regions, and used multiple VLM families to derive pseudo-labels for social activities and inter-person interactions (e.g., sightseeing, drinking, shopping, cooking, moving goods).

Using \dataset{}, we performed a thorough performance evaluation on a suite of state-of-the-art general-purpose VLMs and LLMs on the group detection task. We considered combinations of image, image-and-metadata, and metadata-only variants of these models, \emph{without any task-specific fine-tuning}. The best performing models achieve 66.0 AP (Qwen2.5 VLM~\cite{bai2025qwen2}) and 32.4 F1 score (Gemini 3-Pro). In addition, our results highlight that multiple factors annotated in our benchmark, such as crowd density and world region, significantly affect model performances. For example, models achieved particularly poor performance in underrepresented regions such as Africa and the Middle East, highlighting biases in their general learned representations with respect to specific appearance and action cues of the people and or environment of these regions. For example, our dataset statistics suggest that some actions, such as hand-holding, are far more prevalent in parts of Europe and North America than Africa, which could contribute to this performance discrepancy. Finally, we empirically confirmed that these foundation models without task-specific fine-tuning are still superior to fully supervised baselines on JRDB-Act~\cite{JRDB-Act} that rely on hand-crafted features such as spatial relationships, facial directions, trajectory patterns, and/or individually cropped person features. Hence, while VLM/LLM foundation models are now also the state-of-the-art on social group detection, they still exhibit a clear gap with respect to human perception that warrants further study and exploration.

In summary, this study makes the following contributions: 
(i) we introduce \dataset{}, a first-person view dataset of walking-tour videos from diverse cities that includes dense annotations for social groups and rich scene metadata, enabling realistic social group evaluation across regions, crowd density, and weather conditions;
(ii) we probe a suite of state-of-the-art VLMs and LLMs on social group detection, demonstrating that they surpass fully supervised baselines on JRDB-Act without task-specific fine-tuning;
(iii) we perform a quantitative analysis of crowds and social activities across regions, revealing density and culture-dependent patterns, and highlighting current limits in visual grounding and temporal aggregation for group perception in the wild.
\section{Related Works}
\paragraph{Group Detection.} 
There are several studies on detecting groups from video recordings of moving people. Most cluster trajectories based on simple motion features~\cite{ge2009automatically,khan2015detection,sandikci2011detection,yu2009monitoring,zaidenberg2012generic}. Another line of work focuses on classifying conversational group structures~\cite{inaba2016conversational,thompson2021conversational} by detecting pre-defined constructs, e.g., ``f-formations'' and ``o-structures.'' There are related studies on discovering social roles in groups from images/videos~\cite{ramanathan2013social,solera2013structured}, e.g., parent-child. All of these studies demonstrate their results on a few simple video sequences often shot from an overhead angle (without severe occlusions). In contrast, our goal is to learn subtle patterns from complex, first-person view data, without prior assumptions. 

A few existing works focus on group future state prediction~\cite{alahi2016social,alahi2014socially,bisagno2018group}. They use limited data (two pedestrian datasets), aerial views (avoiding occlusions), non-stochastic future prediction, and do not explicitly take neighboring effects and environment into account. Future prediction of people (not considering group structure)~\cite{afsar2018automatic,bi2020can,hassan2024predicting,hu2020probabilistic,liang2020garden,liang2019peeking} is a more well-studied research area. There are also several studies on detecting group activities~\cite{wu2021comprehensive,zaidenberg2012generic}, and on surveillance and tracking of pedestrians~\cite{bazzani2012decentralized,bazzani2014joint,brunetti2018computer,veeraraghavan2003computer}. However, these works do not make predictions of future states.

Benchmarks in this line are often indoor with controlled recordings with synchronized sensors or outdoor with limited dynamics. For instance, SALSA~\cite{salsa_dataset} offers an indoor dataset containing group social annotations for only two scenarios (poster presentations and cocktail parties), captured using static overhead cameras that oversimplify real-world conditions. Other datasets, including SCU-VSD~\cite{SCU-VSD}, GVEII~\cite{GVEII}, and PANDA~\cite{PANDA}, introduce outdoor scenes but use surveillance or fixed-mount cameras. Recent work introduced more dynamic environment: JLSG~\cite{eccv_groups} annotates social groups in the CUB~\cite{CUB} dataset which has handheld cameras videos, and JRDB-Act~\cite{JRDB-Act} creates their own dataset by traversing indoor and outdoor spaces at Adelaide University with a robot-mounted panoramic camera and provide social grouping and activity labels. However, both remain limited on diverse scenarios and locations. MINGLE~\cite{liu2025minglevlmssemanticallycomplex} offers broader diversity but depends solely on static images and quasi-stationary viewpoints sourced from Google Street View. In contrast, \dataset{} covers a large number of distinct locations and represents our human perspective, emulating day-to-day interactions that more effectively capture real-world dynamics and represent a harder challenge for current models.

A closely related line of work is modeling pedestrian movement. CityWalkers ~\cite{liu2025learningpedgen} is a large-scale dataset of YouTube-curated videos, which leverages pre-trained model-based labeling for individual and scene attributes. This work trains PedGen on CityWalkers to simulate the movement of individuals. Other works ~\cite{bae2022gpgraph, kothari2022traj, gupta2018social} predict pedestrian trajectories, accounting for human-human interactions, and demonstrating efficacy in crowded environments. Introvert ~\cite{shafiee2021introvert} uses conditional visual attention to attend to relevant scene elements. However, these works focus on single pedestrian movements, and do not account for the context of human group activities or group detection.

\paragraph{Cultural benchmarks.} As models become more capable, recent works have started looking at 
cultural benchmarks to tackle the Western-centric trends on LLMs and VLMs. CANDLE~\cite{CANDLE} extracts diverse cultural knowledge from a web corpus, including geography, religion, traditions, and preferences on food, drinks, and clothing. This knowledge serves as a foundation to assess LLMs' cultural awareness. Following work~\cite{culturalVQA, CulturalBench, blendvisbenchmarkingmultimodalcultural}, extend its application to VLMs introducing images, questions and answers with cultural context. Orthogonally, ~\cite{VHELM, VISBIAS} identify biases in models towards an ethnic group. 
ALM-Bench~\cite{all_languages_matter} and CVQA~\cite{CVQA} incorporate different languages to further test cultural awareness in different contexts. 
Additional works test a complex cultural awareness by introducing visual grounding~\cite{satar2025seeingculturebenchmarkvisual, bhatia-etal-2024-local} or text-to-images generation~\cite{jeong-etal-2025-culture}.
This trend highlights the importance and growing interest in evaluating the cultural robustness of large multimodal models; \dataset{} targets first-person view video and social-group structure, offering a complementary lens on how cultural context and crowd dynamics interact in real-world public environments.

\section{\dataset{} Dataset}
We overview \dataset{}, 
a dataset that captures social group behavior from walking-tour videos and aims to better represent social interactions across the globe. To establish a real-world evaluation setting for social group, we annotated our dataset with fine-grained detections and social grouping IDs. This section describes the data collection process (3.1), dataset statistics (3.2), the annotation procedures for coarse and fine-grained group detection (3.3–3.4), and the benchmark metrics (3.5).

\subsection{Data collection}
Our dataset consists of a subset of videos from the Sekai dataset~\cite{li2025sekai}, and a complementary subset of walking-tour videos from YouTube that includes underrepresented cities (e.g., Kenya, Costa Rica, Nepal, etc), totaling over 16 hours. 
We use the official Sekai source code to download a portion of the Sekai-Real-HQ subset. For our complementary collection, we search 
and adapt the Sekai download procedure to avoid overlapping cities. 
We create 5-second clips by sampling from a video at 15-second intervals. We repeat this procedure until the end of each video. For our curated subset, we discard the first two-minute clips, as walking videos often feature highlight moments and frequent scene cuts in their opening segments. 
Following prior work~\cite{li2025sekai}, we use a VLM (i.e., Qwen2.5-VL 72B~\cite{bai2025qwen2}) to perform data curation and cleaning, including discarding clips containing significant scene cuts; categorizing the crowd density, weather, time of day, and indoor/outdoor location. We also leverage country and city information from YouTube metadata. From this 16-hour pool, \dataset{} samples 540 clips (approximately 45 minutes), equally distributed across the three crowd density levels (scattered, moderate, and crowded), comprising 421 clips from the Sekai subset and 119 from our own subset.

\subsection{Dataset Statistics}

\begin{figure}[t!]
\centering
  \includegraphics[width=.90\linewidth]{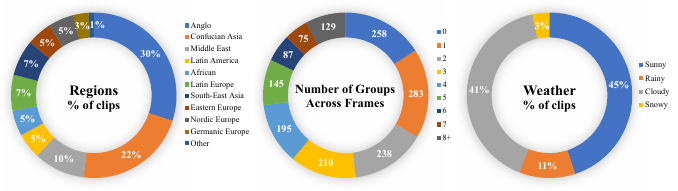}
\caption{Statistics of \dataset{}. Distribution of clips by region (left), number of annotated groups (middle) and by weather (right).}
\label{fig:data_statistics}
\end{figure}

In total, \dataset{} contains clips from 65 countries and 128 cities. Our dataset includes metadata for different levels of crowd density (scattered, moderate, and crowded), time of day (day, night), weather conditions (sunny, cloudy, rainy, and snowy), and an additional distinction between indoor and outdoor environments (see~\autoref{fig:data_statistics} (right) for a detailed percentage distribution of weather). \dataset{} contains 77.96\%  Sekai clips, while the remaining 22.03\% consists of totally new clips from underrepresented cities. ~\autoref{fig:data_statistics} (left), shows the percentage of clips aggregated by region according to the Global Leadership and Organizational Behavior Effectiveness (GLOBE)~\cite{globe, MensahChen2013}. The category “Other” includes countries not covered by this survey, which are often underrepresented, such as Bermuda, Saint Lucia, and the Holy See. A fine-grained count is provided in the supplementary material.

\subsection{Coarse Social Group Annotation}

\begin{figure*}[t!]
    \centering
    
    \begin{subfigure}[t]{0.49\linewidth}
        \centering
        \includegraphics[width=\linewidth]{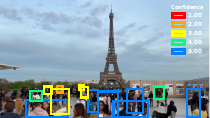}
        \label{fig:annotation_coarse_34}
    \end{subfigure}
    \hfill
    \begin{subfigure}[t]{0.49\linewidth}
        \centering
        \includegraphics[width=\linewidth]{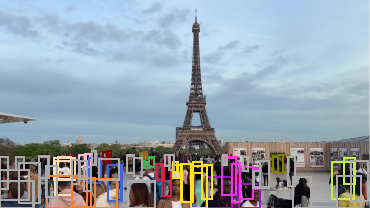}
        \label{fig:annotation_fine_34}
    \end{subfigure}

    \vspace{0.1cm}

    \begin{subfigure}[t]{0.49\linewidth}
        \centering
        \includegraphics[width=\linewidth]{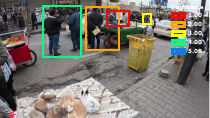}
        \label{fig:annotation_coarse_51}
    \end{subfigure}
    \hfill
    \begin{subfigure}[t]{0.49\linewidth}
        \centering
        \includegraphics[width=\linewidth]{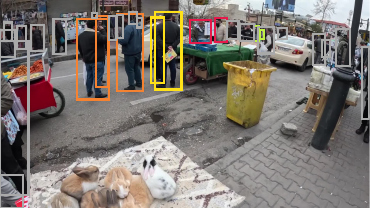}
        \label{fig:annotation_fine_51}
    \end{subfigure}

    \vspace{-0.5cm}
    \caption{\textbf{Annotation samples in \dataset{}.} We show group detections with varying annotator confidence levels (red = lowest, blue = highest) and corresponding fine-grained annotations (individuals marked in gray).}
    \vspace{-0.2cm}
    \label{fig:annotation_examples}
\end{figure*}

To obtain ground-truth labels for social groups, we manually annotate \dataset{}.  We design a web-based interface for the annotation process, requiring annotators to draw bounding boxes and assign a confidence score on a Likert scale (1 to 5, where higher is more confident) to groups detected in a set of representative frames temporally distributed across each clip. Annotators also had the choice to view the full clip to resolve ambiguous cases. We organized the 540 clips into 36 tasks of 15 clips each with a specific frame, and distributed the annotation tool. Each frame has at least 3 distinct annotators. Finally, to create a single annotation set, we iterate through each detected group and add it only if it does not overlap (IoU $\geq$ 0.3) with a previously added bounding box. We follow this approach to maximize group detections, as individual annotators often miss some groups. ~\autoref{fig:data_statistics} (middle) presents the aggregated number of groups annotated across frames for our coarse group evaluation task, with 
at least two groups for most of the images, and a few with more than 8 annotations. 
\autoref{fig:annotation_examples} provides examples illustrating different confidence levels for this grouping task. 
In particular, there are some frames with up to 26 groups in a single frame.


\subsection{Fine-grained Social Group Annotation}

First, we generate pseudo fine-grained annotations using SAM3~\cite{carion2025sam3segmentconcepts} detections; a step-by-step guide is provided in the supplementary material. If the center point of a detection lies within a coarse patch, its label is assigned as the smallest enclosing coarse annotation. To further refine these detections, we deploy a web-based interface where annotators can reassign the label of a detection, remove a detection or an entire group, and add a bounding box. On average, each annotated frame is reviewed by approximately three annotators, with an inter-annotator agreement of 95.56\% with the preceding coarse annotations. Finally, to consolidate the annotations, we apply a majority-vote scheme with a connected-components heuristic: each group label must appear in at least half of the annotators, and candidate boxes across annotators are matched with an area-adaptive IoU. Only connected components with at least two members are retained and merged to produce the final fine-grained detections, yielding a fine-grained 
inter-annotator 
agreement score of 91.64\%. Totally, we have 24.331 annotated bounding boxes and 5.151 groups. \autoref{fig:annotation_examples} provides examples of the fine-grained annotations along with their 
initial candidate 
coarse annotation.

\subsection{Social Group Benchmark and Metrics}

We use the JRDB-Act~\cite{JRDB-Act} dataset and \dataset{} as a benchmark for social group detection. 
Following prior works~\cite{Yokoyama_2025_ICCV, PANDA, JRDB-Act, 10286297}, 
we use the annotated bounding boxes for evaluation and the following metrics. We evaluate social grouping with social average precision ($G_{1}$-$G_{5}$, AP) metric used in JRDB-Act~\cite{JRDB-Act}. For each group, we establish correspondences between predicted and ground-truth groups by solving an optimal assignment problem, count true positives as matched boxes with IoU~$\geq 0.5$, and compute average precision over all confidence thresholds. In addition, we evaluate the grouping IDs. Given the ground-truth ($G_{\text{gt}}$) and predicted ($G_{\text{det}}$) sets of persons, we consider $G_{\text{det}}$ a true positive if $\frac{|G_{\text{det}} \cap G_{\text{gt}}|}{\max(|G_{\text{det}}|, |G_{\text{gt}}|)} > 0.5$ and a false positive otherwise. Finally, we compute precision, recall, and F1.

\begin{figure}[h!] 
    \centering
    \includegraphics[width=1\linewidth]{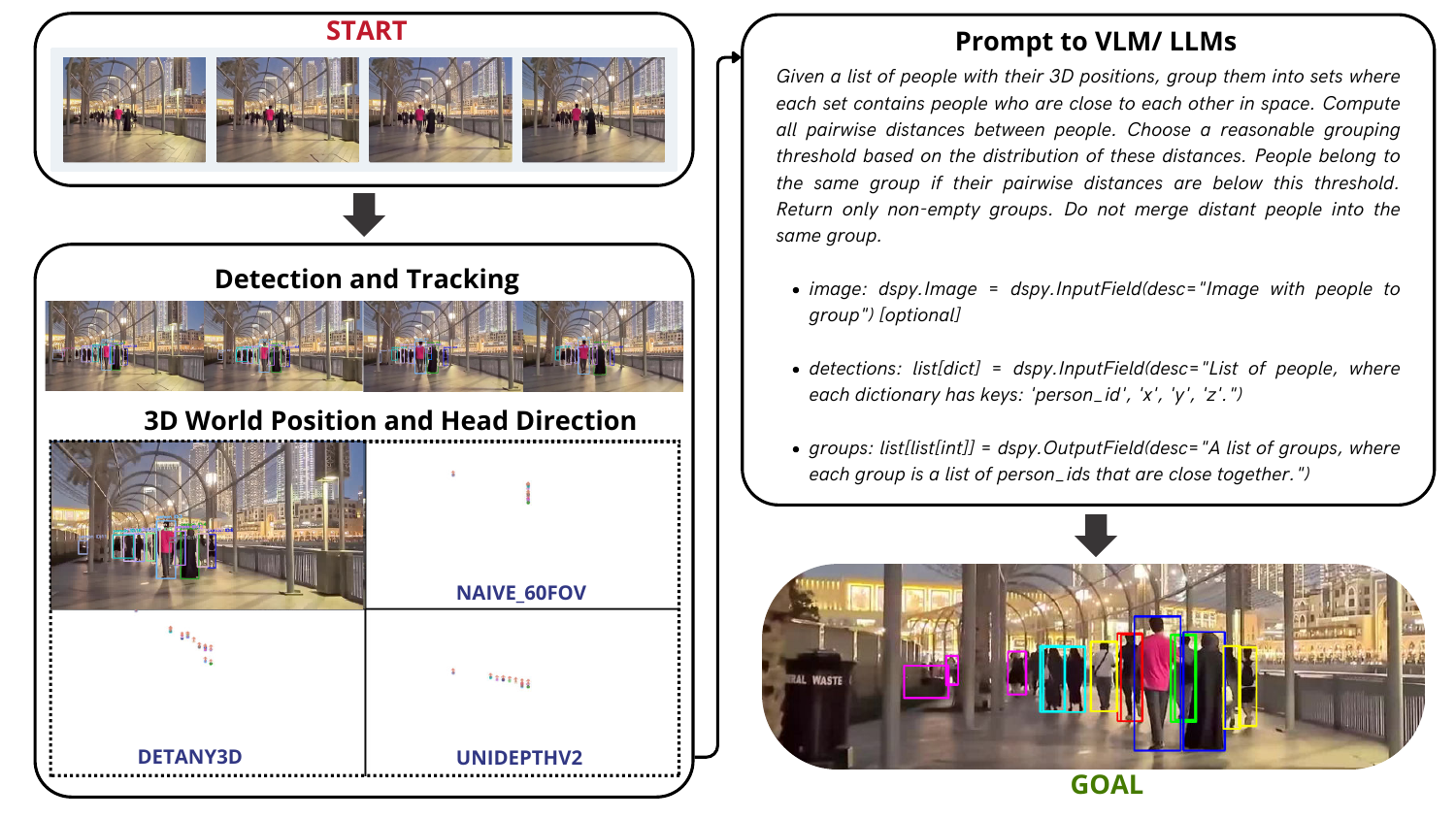}
    \caption{\textbf{Overview of the pipeline for social grouping}. Our process involves bounding along with track\_ID to generate 3D metadata for an effective prompting. 
    }
    \label{fig:approach}
\end{figure}

\section{Detecting Social Groups in the Wild}
\label{sec:det_in_the_wild}

We implement a pipeline to obtain diverse metadata from bounding box within track ids (4.1) and develop an effective prompting strategy for LLMs and VLMs to detect groups in first-person view videos (4.2).
The overall procedure is illustrated in \autoref{fig:approach}. 

\subsection{3D metadata generation}
\label{sec:3d_pos}

Given a set of $\mathbf{P}$ persons, where each $\mathbf{p_i}$ consists of a track ID $\mathbf{t}_i$ and a bounding box $\mathbf{b}_i = (x_1, y_1, x_2, y_2)$ in the standard top-left and bottom-right format. For each person $\mathbf{p_i}$, we estimate their corresponding 3D world position to obtain depth-related information. This is a crucial and challenging step, as intrinsic and extrinsic camera parameters are often unavailable. 
We adopt two approaches:  

\begin{itemize}
    \item \textbf{Estimated intrinsics approach:} We assume an intrinsic camera ($\mathbf{K}$) obtained from an estimated field of view (FOV). The optical center is obtained with $c_x = W/2, c_y = H/2 $, where W and H are the width and the height of the frame dimensions, respectively. Then, the focal length $f_x = \frac{W/2}{\tan(\mathrm{FOV}_x / 2)}$, where $\mathbf{\mathrm{FOV}_x}$ is the field of view, we do similarly to obtain $f_y$. Finally, we compute 3D coordinate correspondences using classical 2D-to-3D projection formulas. For more details, ablation and full comparison with other approaches are presented in the supplementary material. 
    \item \textbf{Off-the-shelf model approach:} We use pretrained models such as UniDepth-V2~\cite{piccinelli2025unidepthv2} or DetAny3D~\cite{zhang2025detect}. Specifically, we provide each 2D bounding box to DetAny3D and query the center of the bbox to UniDepthV2 and retrieve the world coordinates from its prediction.  

\end{itemize}

Finally, we aggregate $\mathbf{w}_i = (x_i,y_i,z_i) $ for each person $\mathbf{p}_i$. 

\subsection{Prompt Design}
\label{sec:prompt_design}

We design specific prompt to query VLMs and LLMs. 
We use dspy~\cite{khattab2024dspy} to standardize query generation and evaluation across different models, ensuring consistent comparisons. The model receives each person track\_ID with its corresponding 3D positions within a prompt to evoke group reasoning. For VLM, model receive the image alongside visual hints that link each track\_ID to the corresponding person. Prompt details are available in the appendix.

\section{Experimental Results}

\begin{table*}[ht!]
    \centering
    \footnotesize
    \caption{Fine-grained performance comparison on \dataset{} dataset segmented by crowd density: scattered, moderate and crowded. Model combination indicates the backbone family, number of parameters, and modality. $G_1$--$G_5$ represent AP scores for group sizes 1--5$^+$, with AP showing the average performance across all groups.} 
    \vspace{0.3cm}
    \resizebox{0.99\linewidth}{!}{
    \begin{tabular}{lllccccccccccccccccccc}
    \toprule
        \multirow{2}{*}{Model} &  & 
        & \multicolumn{6}{c}{\textbf{Scattered}}
        & \multicolumn{6}{c}{\textbf{Moderate}} & \multicolumn{6}{c}{\textbf{Crowded}} & \textbf{All} \\
        \cmidrule(lr){4-9} \cmidrule(lr){10-15} \cmidrule(lr){16-21} \cmidrule(lr){22-22}
        & & & G$_{1}$ & G$_{2}$ & G$_{3}$ & G$_{4}$ & G$_{5}$ & AP & G$_{1}$ & G$_{2}$ & G$_{3}$ & G$_{4}$ & G$_{5}$ & AP & G$_{1}$ & G$_{2}$ & G$_{3}$ & G$_{4}$ & G$_{5}$ & AP & AP\\
    \midrule
        \multirow{2}{*}{\rotatebox{10}{\shortstack{Cosmos-\\Reason2}}} & \multirow{2}{*}{8B} & \multirow{1}{*}{VLM} & 57.18 & 74.42 & 76.36 & 75.35 & 80.6 & 72.78 & 28.92 & 45.28 & 59.54 & 68.02 & 68.55 & 54.06 & 12.8 & 30.12 & 50.05 & 61.53 & 73.18 & 45.53 & 51.07 \\
         &  & \multirow{1}{*}{LLM} & 60.72 & 71.03 & 72.84 & 70.49 & 75.0 & 70.02 & 32.47 & 49.61 & 65.81 & 62.54 & 76.02 & 57.29 &  20.58 & 40.56 & 55.9 & 59.69 & 65.15 & 48.38 & 53.38 \\
         \midrule
        \multirow{4}{*}{\rotatebox{10}{\shortstack{Qwen2.5}}} & \multirow{2}{*}{32B} & \multirow{1}{*}{VLM} & 58.94 & 83.37 & \textbf{86.92} & \textbf{81.6} & \textbf{92.27} & \textbf{80.62} & 30.28 & 62.72 & 77.55 & \textbf{84.21} & 79.38 & 66.83 & 19.11 & 52.42 & 73.26 & 76.05 & 71.56 & 58.48 & 63.37\\
         &  & \multirow{1}{*}{LLM} & 63.71 & 85.28 & 82.38 & 79.51 & 62.73 & 74.72 & 40.49 & 73.14 & 75.28 & 73.02 & 70.62 & 66.51 & 30.45 & 64.66 & 72.69 & 68.27 & 64.38 & 60.09 & 63.54 \\
        \cmidrule(lr){3-22}
         & \multirow{2}{*}{72B} & \multirow{1}{*}{VLM}  & 60.53 & 85.1 & 85.11 & 79.17 & 84.54 & 78.89 &  37.18 & 66.91 & \textbf{79.14} & 78.45 & \textbf{82.76} & \textbf{68.89} & 25.89 & 59.79 & \textbf{73.79} & \textbf{77.81} & \textbf{74.18} & \textbf{62.29} & \textbf{66.00} \\
         &  & \multirow{1}{*}{LLM}   & 63.97 & \textbf{85.84} & 80.24 & 75.0 & 59.55 & 72.92 & 38.73 & 73.68 & 76.4 & 71.85 & 69.28 & 65.99 & 28.02 & 65.93 & 73.18 & 68.64 & 61.92 & 59.54 & 62.93 \\
         \midrule
        \multirow{2}{*}{\rotatebox{10}{\shortstack{Qwen3}}} & \multirow{2}{*}{30B} & \multirow{1}{*}{VLM} & 71.59 & 80.28 & 74.26 & 75.0 & 59.09 & 72.04 & 50.29 & 66.02 & 71.33 & 71.07 & 70.36 & 65.81 & 36.89 & 57.26 & 63.7 & 60.82 & 58.39 & 55.41 & 63.23 \\
         &  & \multirow{1}{*}{LLM} & 69.41 & 83.56 & 79.37 & 78.47 & 62.73 & 74.71 & 54.94 & 69.94 & 71.17 & 72.14 & 70.15 & 67.67 &  47.59 & 62.84 & 69.35 & 63.25 & 59.25 & 60.46 & 64.23 \\
         \midrule
        \multirow{2}{*}{\rotatebox{10}{\shortstack{Gemini-\\3-Pro}}} &  & \multirow{1}{*}{VLM}  & \textbf{82.67} & 80.85 & 73.92 & 76.39 & 68.79 & 76.52 & \textbf{70.37} & \textbf{75.5} & 68.24 & 67.24 & 57.07 & 67.69 & \textbf{63.09} & \textbf{69.78} & 64.16 & 54.43 & 47.9 & 59.87 &  64.05 \\
         &  & \multirow{1}{*}{LLM} & 77.66 & 81.85 & 75.33 & 75.35 & 59.55 & 73.95 & 60.29 & 74.18 & 73.23 & 70.9 & 62.12 & 68.14 & 50.95 & 69.0 & 68.63 & 63.71 & 54.88 & 61.44 & 64.85 \\
    \bottomrule
    \end{tabular}
    }
    \label{table:results_segmented_ours_AP}
\end{table*}

\paragraph{Detecting groups on \dataset{} benchmark} We conduct experiments
across one propietary and three open-source models: Qwen2.5, Qwen3, Cosmos-Reason2 and Gemini 3. We test both VLM and LLM counterparts for each model, covering 8B model for Cosmos-Reason2, 32B and 72B for Qwen2.5, 30B for Qwen3, as well as the Pro version for Gemini-3. VLM models receive an image and a prompt as input, whereas LLMs receive only a textual prompt. We evaluate VLM/LLMs using our fine-grained annotations and metrics across region and crowdness dimensions. The results are presented in \autoref{table:results_segmented_ours_AP}, \ref{table:results_segmented_ours_F1} and \ref{table:results_regions_ours_AP}. In particular, \autoref{table:results_segmented_ours_AP} and \ref{table:results_segmented_ours_F1} report the AP and F1 metric under varying crowd densities alongside with the overall performance. \autoref{table:results_regions_ours_AP} shows AP results segmented by the GLOBE regions, highlighting the best (in red) and worst performance (in blue) row-wise.

\begin{table*}[bt!]
    \centering
    \footnotesize
    \caption{Fine-grained performance comparison on \dataset{} dataset segmented by crowd density: scattered, moderate and crowded. Model combination indicates the backbone family, number of parameters, and modality. Precision, Recall, and F1 scores measure the false positive rate of predicted group IDs (without individuals).} 
    \resizebox{0.85\linewidth}{!}{
    \begin{tabular}{lllcccccccccc}
    \toprule
        \multirow{2}{*}{Model} & & 
        & \multicolumn{3}{c}{\textbf{Scattered}}
        & \multicolumn{3}{c}{\textbf{Moderate}} & \multicolumn{3}{c}{\textbf{Crowded}} & \textbf{All}\\
        \cmidrule(lr){4-6} \cmidrule(lr){7-9} \cmidrule(lr){10-12} \cmidrule(lr){13-13}
        & & & Precision & Recall & F1 & Precision & Recall & F1   & Precision & Recall & F1 & F1 \\
    \midrule
        \multirow{2}{*}{\rotatebox{10}{\shortstack{Cosmos-\\Reason2}}} & \multirow{2}{*}{8B} & \multirow{1}{*}{VLM} & 50.44 & 49.82 & 50.1 & 25.92 & 16.16 & 19.9 &  11.22 & 5.41 & 7.3 & 17.64 \\
         &  & \multirow{1}{*}{LLM} & 44.73 & 40.6 & 42.47 &  22.04 & 14.3 & 17.33 & 11.75 & 6.61 & 8.45 &  15.93 \\
        \midrule
        \multirow{4}{*}{\rotatebox{10}{\shortstack{Qwen2.5}}} & \multirow{2}{*}{32B} & \multirow{1}{*}{VLM} & 51.4 & 61.87 & 56.12 &  27.05 & 26.18 & 26.58 &  14.35 & 13.69 & 14.01 & 23.49 \\
         &  & \multirow{1}{*}{LLM}  & 51.22 & 64.05 & 56.88 & 29.8 & 33.25 & 31.41 & 17.5 & 21.74 & 19.38 & 27.70 \\
        \cmidrule(lr){3-13}
         & \multirow{2}{*}{72B} & \multirow{1}{*}{VLM} & 52.77 & \textbf{65.65} & 58.49 & 32.76 & 33.54 & 33.1 &  21.0 & 22.25 & 21.6 & 29.97 \\
         &  & \multirow{1}{*}{LLM}  & 51.09 & 65.36 & 57.3 & 28.52 & 32.39 & 30.31 &  15.01 & 18.98 & 16.75  & 25.97 \\
        \midrule
        \multirow{2}{*}{\rotatebox{10}{\shortstack{Qwen3}}} & \multirow{2}{*}{30B} & \multirow{1}{*}{VLM} & 54.88 & 56.73 & 55.7 & 30.02 & 25.69 & 27.68 & 13.73 & 12.45 & 13.06 & 23.11 \\
         &  & \multirow{1}{*}{LLM} & 55.96 & 61.84 & 58.73 & 33.72 & 29.54 & 31.47 & 17.92 & 16.29 & 17.03 & 27.05 \\
         \midrule
        \multirow{2}{*}{\rotatebox{10}{\shortstack{Gemini-\\3-Pro}}} &  & \multirow{1}{*}{VLM} & \textbf{61.02} & 57.83 & \textbf{59.19} & \textbf{38.44} & \textbf{36.99} & \textbf{37.63} & \textbf{22.75} & \textbf{25.67} & \textbf{24.09} & \textbf{32.44} \\
         &  & \multirow{1}{*}{LLM} & 58.0 & 59.01 & 58.26 & 34.59 & 34.21 & 34.34 & 19.28 & 22.72 & 20.83 & 29.41
         \\
    \bottomrule
    \end{tabular}
    }
    \label{table:results_segmented_ours_F1}
\end{table*}

\begin{table*}[h!]
    \centering
    \footnotesize
    \caption{Fine-grained performance comparison on the \dataset{} dataset, segmented by GLOBE regions: African (AF), Anglo (AN), Confucian Asia (CA), Eastern Europe (EU), Germanic Europe (GE), Latin America (LA), Latin Europe (LE), Middle East (ME), Nordic Europe (NE), South-East Asia (SA), Other (O). Model combination indicates the backbone family, number of parameters, and input modality. We report each region's AP score individually, and compute MAP as the average across all regions. The best and worst values per row are highlighted in red and blue, respectively.}  
    \vspace{0.3cm}
    \resizebox{0.8\linewidth}{!}{
    \begin{tabular}{lllccccccccccc}
    \toprule
        \multirow{1}{*}{Model} &  &  
        & \textbf{AF}
        & \textbf{AN} & \textbf{CA} & \textbf{EU} & \textbf{GE} & \textbf{LA} & \textbf{LE} & \textbf{ME} &\textbf{NE} & \textbf{SA} & \textbf{O}  \\
    \midrule
        \multirow{2}{*}{\rotatebox{10}{\shortstack{Cosmos-\\Reason2}}} & \multirow{2}{*}{8B} & \multirow{1}{*}{VLM} & 47.19 & 50.49 & 54.01 & 51.69 & \blue{40.07} & 51.12 & 52.98 & 47.05 & 49.86 & 52.86 & \red{58.89} \\
         &  & \multirow{1}{*}{LLM}  & \blue{45.95} & \red{56.89} & 55.25 & 54.59 & 50.95 & 54.17 & 54.40 & 49.18 & 51.21 & 52.95 & 53.37 \\
         \midrule
        \multirow{4}{*}{\rotatebox{10}{\shortstack{Qwen2.5}}} & \multirow{2}{*}{32B} & \multirow{1}{*}{VLM}& 61.99 & 64.63 & 63.57 & 58.82 & 60.48 & 66.55 & 64.09 & 57.98 & \red{71.75} & 65.75 & \blue{50.00} \\
         &  & \multirow{1}{*}{LLM}  & 61.87 & 65.01 & 62.16 & 61.69 & 62.84 & \red{66.89} & 63.84 & \blue{59.75} & 69.36 & 63.28 & 61.63  \\
        \cmidrule(lr){3-14}
         & \multirow{2}{*}{72B} & \multirow{1}{*}{VLM} & \blue{58.64} & 67.24 & 65.78 & 64.54 & 67.06 & \red{70.06} & 67.69 & 59.42 & 71.45 & 67.30 & 65.80 \\
         &  & \multirow{1}{*}{LLM}  & 61.94 & 65.22 & 61.88 & 62.32 & 64.28 & 65.04 & 62.61 & 57.33 & \red{67.31} & 64.49 & \blue{54.21} \\
         \midrule
        \multirow{2}{*}{\rotatebox{10}{\shortstack{Qwen3}}} & \multirow{2}{*}{30B} & \multirow{1}{*}{VLM} & 56.07 & 64.76 & 58.49 & 60.26 & \red{65.85} & 59.92 & 63.42 & \blue{53.94} & 56.30 & 64.20 & 63.10  \\
         &  & \multirow{1}{*}{LLM}  & 62.64 & 65.61 & 64.43 & 65.95 & 66.34 & 66.73 & 63.30 & 57.70 & \red{67.66} & 66.31 & \blue{55.60}  \\
         \midrule
        \multirow{2}{*}{\rotatebox{10}{\shortstack{Gemini-\\3-Pro}}} &  & \multirow{1}{*}{VLM}  & 62.41 & 66.70 & 64.01 & 68.23 & 65.53 & 64.79 & 64.32 & \blue{60.24} & \red{71.84} & 66.03 & 62.31 \\
         &  & \multirow{1}{*}{LLM} & 60.87 & 64.59 & 64.69 & \red{68.61} & 64.37 & 62.55 & 66.47 & \blue{59.51} & 62.44 & 66.69 & 64.70 \\
    \bottomrule
    \end{tabular}   
    }
\label{table:results_regions_ours_AP}
\end{table*}

\begin{table*}[h!]
    \centering
    \footnotesize
    \caption{Fine-grained performance comparison on the JRDB dataset. Model combination indicates the backbone family, number of parameters, and input modality. $G_1$--$G_5$ represent AP scores for group sizes 1--5$^+$, with AP showing the average performance across all groups. Precision, Recall, and F1 scores measure the false positive rate of predicted group IDs (without individuals).} 
    \resizebox{0.8\linewidth}{!}{
    \begin{tabular}{lllccccccccc}
    \toprule
        \multirow{2}{*}{Model} & & &
        \multicolumn{6}{c}{\textbf{Group Detection (AP)}} &
        \multicolumn{3}{c}{\textbf{Group ID Prediction}} \\
        \cmidrule(lr){4-9} \cmidrule(lr){10-12}
        & & & G$_{1}$ & G$_{2}$ & G$_{3}$ & G$_{4}$ & G$_{5}$ & AP  & Precision & Recall & F1 \\
    \midrule
    \rowcolor{gray!20}
         \multicolumn{2}{l}{JLSG~\cite{eccv_groups}} & & 8.0 & 29.30 & 37.5 & 65.40 & 67.0 & 41.4 & - & - & - \\
    \rowcolor{gray!20}
         \multicolumn{2}{l}{JRDB-Act~\cite{JRDB-Act}} & & 81.40 & 64.80 & 49.10 & 63.20 & 37.20 & 59.2 & - & - & - \\
    \rowcolor{gray!20}              \multicolumn{2}{l}{DVT3~\cite{Yokoyama_2025_ICCV}} &    & - & - & - & - & - & - &  \textbf{61.16} & 31.06 & 41.19 \\
    \midrule
        \multirow{2}{*}{\rotatebox{10}{\shortstack{Cosmos-\\Reason2}}}
         & \multirow{2}{*}{8B} & \multirow{1}{*}{VLM} &
          44.88 & 34.7 & 30.96 & 36.83 & 55.0 & 40.47 & 19.83 & 5.83 & 9.02 \\
         &  & \multirow{1}{*}{LLM} & 48.47 & 36.19 & 29.05 & 31.71 & 56.26 & 40.34 & 24.7 & 6.93 & 10.82 \\
        \midrule
        \multirow{4}{*}{\rotatebox{10}{\shortstack{Qwen2.5}}}
         & \multirow{2}{*}{32B} & \multirow{1}{*}{VLM} &
          25.54 & 47.44 & 47.23 & 54.03 & 69.3 & 48.71 & 22.39 & 16.81 & 19.2 \\
         &  & \multirow{1}{*}{LLM} & 35.13 & 56.78 & 54.11 & 56.96 & 69.49 & 54.49 & 28.45 & 23.88 & 25.97 \\
        \cmidrule(lr){2-12}
         & \multirow{2}{*}{72B} & \multirow{1}{*}{VLM} &
          28.4 & 45.44 & 43.62 & 51.26 & 71.99 & 48.14 & 23.54 & 17.28 & 19.93 \\
         & & \multirow{1}{*}{LLM} & 32.95 & 55.42 & 55.34 & 60.99 & 66.86 & 54.31 & 28.98 & 25.46 & 27.11 \\
        \midrule
        \multirow{4}{*}{\rotatebox{10}{Qwen3}}
         & \multirow{2}{*}{30B} & \multirow{1}{*}{VLM} &
         30.5 & 44.2 & 41.5 & 44.71 & 72.48 & 46.68 & 30.19 & 17.45 & 22.12 \\
         &  & \multirow{1}{*}{LLM} & 47.15 & 57.38 & 54.45 & 57.97 & 66.62 & 56.71 & 40.25 & 27.84 & 32.91 \\
        \cmidrule(lr){2-12}
         & \multirow{2}{*}{235B} & \multirow{1}{*}{VLM} & 31.54 & 44.28 & 41.36 & 50.59 & 77.24 & 49.0 & 36.80 & 21.51 & 27.15 \\
         &  & \multirow{1}{*}{LLM} & 31.0 & 51.29 & 60.4 & 69.63 & \textbf{79.95} & 58.46 & 34.03 & 25.53 & 29.17 \\
        \midrule
        \multirow{2}{*}{\rotatebox{10}{\shortstack{Gemini-\\3-Pro}}} &
        & \multirow{1}{*}{VLM} & \textbf{56.66} & \textbf{71.79} & 79.33 & 79.7 & 57.49 & 68.99 & 42.99 & \textbf{46.38} & \textbf{44.62} \\
        &  & \multirow{1}{*}{LLM} & 51.65 & 69.97 & \textbf{82.52} & \textbf{81.8} & 64.08 & \textbf{70.0} & 40.89 & 43.97 & 42.37 \\
    \bottomrule
    \end{tabular}
    }
    \label{table:results_JRDB_singleframe}
\end{table*}

\paragraph{Detecting Groups from Panoramic Views.}

\autoref{table:results_JRDB_singleframe} shows experimental results on the JRDB-Act datasets. Performance is evaluated using fine grained Average Precision (AP), following the same setup as ~\cite{JRDB-Act} across five group-size categories: single groups (G$_1$), two groups (G$_2$), three groups (G$_3$), four groups (G$_4$), and five or more groups (G$_5$) as well the overall mean AP (AP). Along with these metrics, we present Precision, Recall and F1 score for groups excluding individuals. These results includes trained methods within this dataset.

\paragraph{Detecting Social Activities on \dataset{}.}

\begin{figure}[h!] 
    \centering
    \includegraphics[width=1\linewidth]{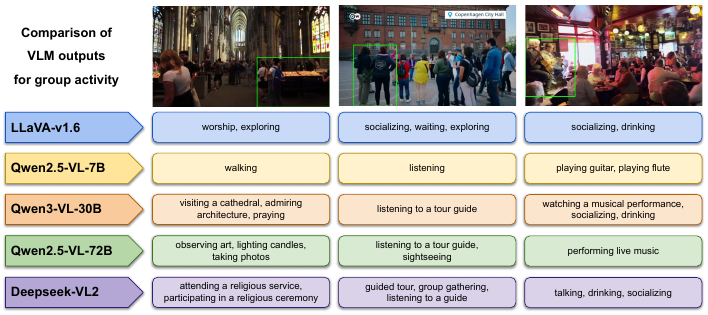}
    \caption{Qualitative comparison of group activities predicted by VLMs for various scenarios.}
    \label{fig:cultural_samples}
\end{figure}

\begin{figure}[h!] 
    \centering
    \includegraphics[width=1\linewidth]{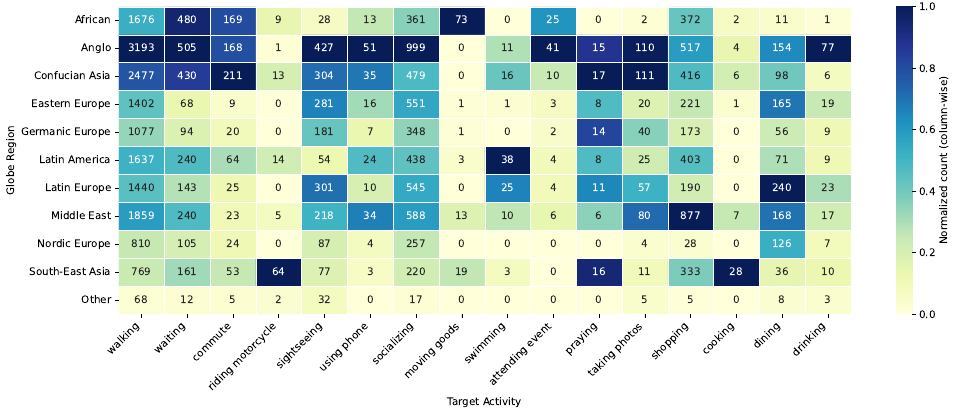}
    \vspace{-0.7cm}
    \caption{Per-region activity count heatmap across GLOBE region by Qwen2.5-VL 72B Instruct. Column-wise color intensity reflects the relative frequency of each activity within each region.}
\label{fig:activity_count_by_region}
\end{figure}

Moreover, we probe different VLM models to analyze activities and close interactions in the detected group. For each frame, we compute a bounding box for each group ($B_\text{group}$) encompassing its members.
We prompt different model versions of Qwen2.5-VL with the respective frame and $B_\text{group}$ to describe activities in which members of a group are engaged, and answer whether or not there are close interactions within a group, particularly probing for hugging and hand-holding (results in the appendix).
Activity descriptions are left open-ended, which leads to outcomes that are either verbs (e.g. \textit{walking}, \textit{socializing}) or longer descriptions such as \textit{waiting to cross the street in snowy weather}. 
We prompt different VLM families (i.e., LLaVA-1.6, Qwen2.5-VL and DeepSeek-VL2) for this task. \autoref{fig:cultural_samples} presents an example of activities across models. To enable further analysis with activity frequencies, we restrict the open-ended outputs to a fixed vocabulary by deduplicating semantically similar words and descriptions (e.g., \textit{taking a stroll} and \textit{walking in the city} map to \textit{walk}). For this, we convert unique activity values to an embedding space using Qwen3-8B-Embedding and perform greedy matching based on cosine similarity. We then compute the frequencies of the activities in the vocabulary. \autoref{fig:activity_count_by_region} shows count of clustered activities per region by Qwen2.5 VL 72B Instruct.

\section{Analysis}

\begin{figure*}[h!]
    \centering
    \begin{subfigure}[b]{0.32\linewidth}
        \centering
        \includegraphics[width=\linewidth]{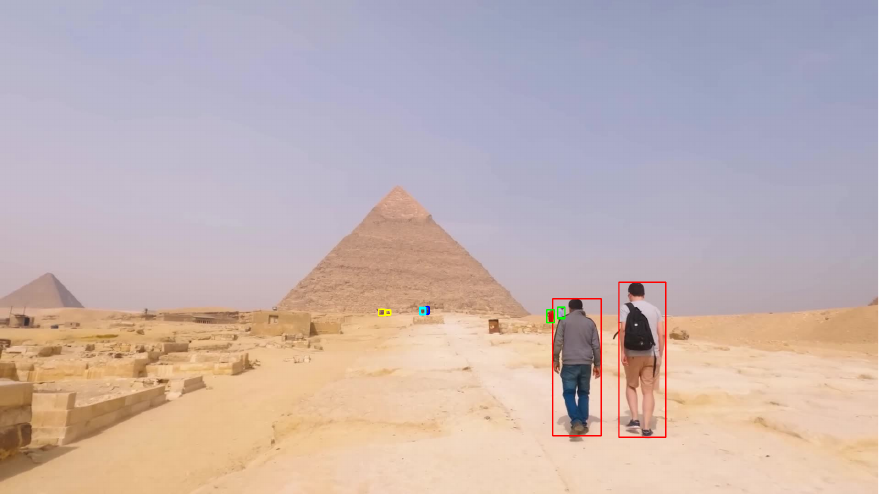}
        \label{fig:annotation_a1}
    \end{subfigure}
    \hfill
    \begin{subfigure}[b]{0.32\linewidth}
        \centering
        \includegraphics[width=\linewidth]{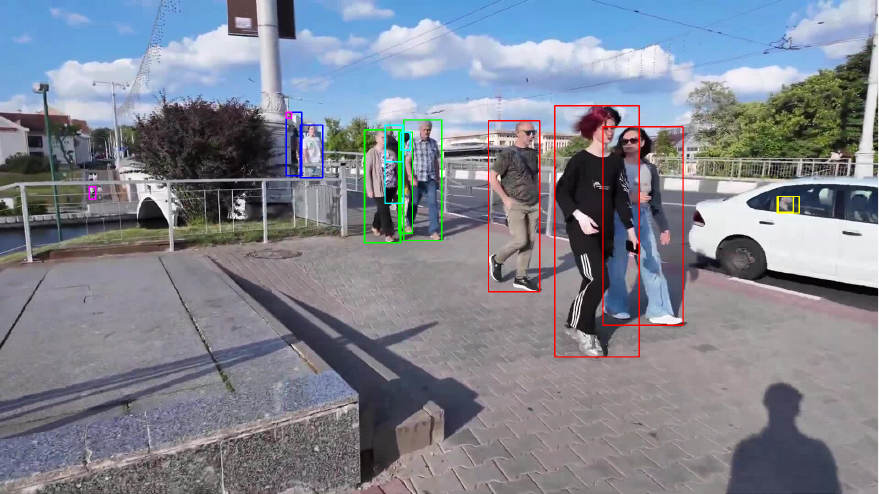}
        \label{fig:annotation_b1}
    \end{subfigure}
    \hfill
    \begin{subfigure}[b]{0.32\linewidth}
        \centering
        \includegraphics[width=\linewidth]{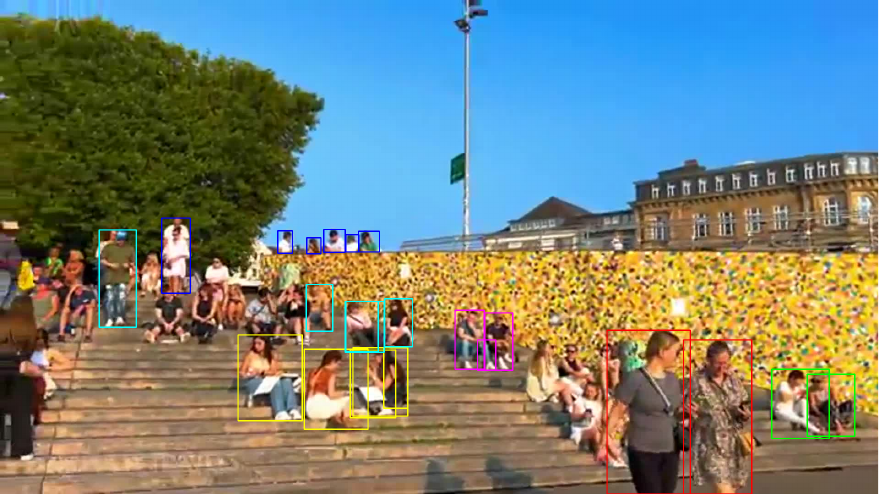}
        \label{fig:annotation_c1}
    \end{subfigure}

    \vspace{0.1cm}

    \begin{subfigure}[b]{0.32\linewidth}
        \centering
        \includegraphics[width=\linewidth]{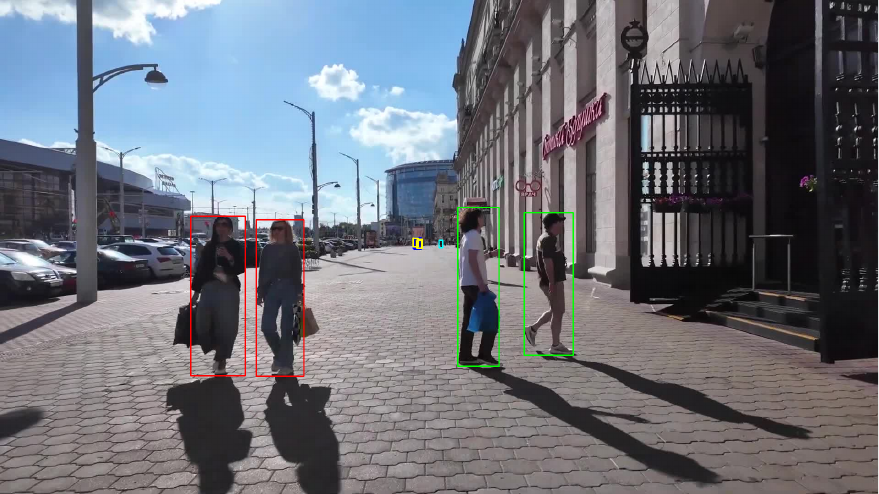}
        \caption{Scattered}
        \label{fig:annotation_a2}
    \end{subfigure}
    \hfill
    \begin{subfigure}[b]{0.32\linewidth}
        \centering
        \includegraphics[width=\linewidth]{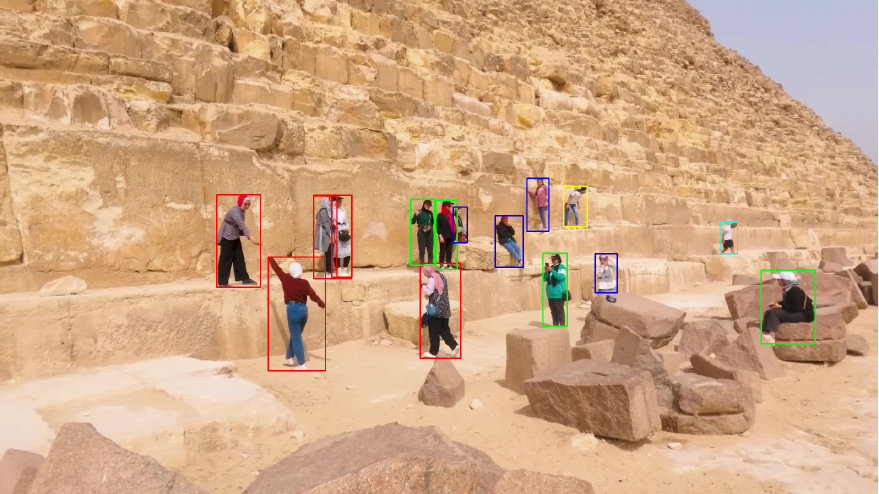}
        \caption{Moderate}
        \label{fig:annotation_b2}
    \end{subfigure}
    \hfill
    \begin{subfigure}[b]{0.32\linewidth}
        \centering
        \includegraphics[width=\linewidth]{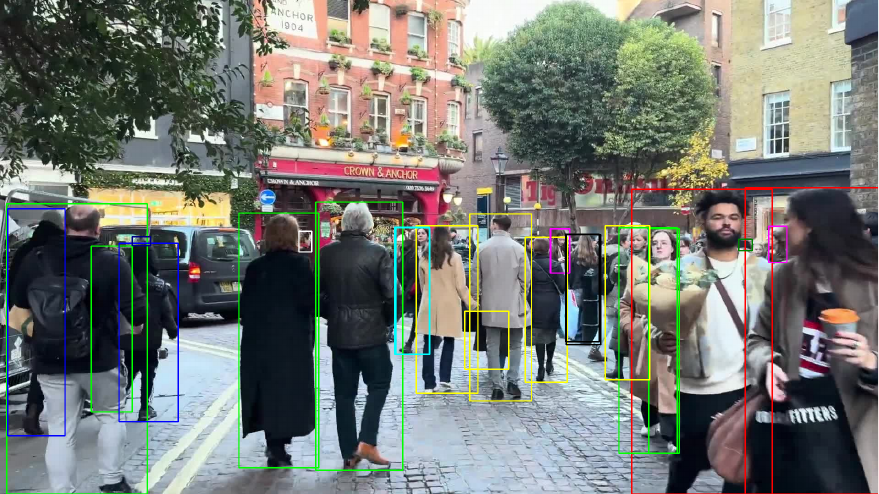}
        \caption{Crowded}
        \label{fig:annotation_c2}
    \end{subfigure}

    \caption{Qualitative results of our approach across frames with varying crowd densities. From left to right: scattered, moderate, and crowded. Detections with the same color are judged to be in the same social group.}
    \label{fig:qualitative_results}
\end{figure*}

\paragraph{Social Grouping.}
On \dataset{}, we observe that larger models generally achieve higher AP, with Qwen 2.5 72B, 32B, and Gemini 3-Pro yielding the best performance. In particular, Qwen2.5VL 72B achieves 66.00 overall AP, while Gemini 3-Pro achieves 32.44 F1, outperforming all LLM variants of comparable size. When segmented by crowd density, the results reveal an expected pattern: as crowd density increases, the task becomes increasingly challenging for all models, yielding consistently lower scores in both metrics. For instance, best performance on scattered set drops from 80.62\% to 62.29\% in AP and from 59.19\% to 32.44\% in F1 score. Qualitative results in \autoref{fig:qualitative_results} illustrate that VLM/LLMs can recover reasonable groupings in scattered and moderate scenes, but still struggle in the most crowded scenarios. Region-wise results reveal significant performance variation across GLOBE regions. Middle East (ME) and African (AF) regions consistently record the lowest scores, whereas Nordic Europe (NE), Anglo (AN), and Latin America (LA) prove less challenging for all models. Interestingly, the better-performing regions are largely or partially influenced by Western culture, suggesting that this disparity may reflect inherent biases in the models, where Western social contexts and interaction patterns are better represented.

On the JRDB-Act benchmark,  
smaller models tends to be better in groups $G_{1}$ and $G_{2}$, and $G_{5}$ are best in larger models. Gemini 3-Pro reaches 70.00 on AP and 44.62 on F1, outperforming the supervised baselines by +10.8 and +3.43 respectively. 
This suggests that current VLMs/LLMs can surpass trained models when perceptual metadata is available. However, some challenges remain; compared with DVT3~\cite{Yokoyama_2025_ICCV}, VLMs and LLMs achieve better recall but lower precision, suggesting a tendency to over-predict group memberships, capturing more true groups at the cost of introducing false positives. In this dataset, LLMs consistently surpass VLMs of similar size when both are given access to structured metadata, which may be attributed to the panoramic image format.

\paragraph{Activity Recognition.}
Based on activity recognition results, we observe that certain actions exhibit a potential relation to a geographic context. For instance, ``drinking'' appears as a top activity in Latin Europe, Anglo, and the Middle East, while ``moving goods'' is prominent in Africa and ``cooking'' in South East Asia.
Beyond the most frequent actions related to walking, waiting, and socializing, the most frequent activities across models are: ``moving goods'' for Africa, ``riding a motorcycle'' for Asia.
These patterns may reflect genuine behavioral differences captured in the first-person view videos; however, they could also reveal systematic biases in the training data or inherent cultural stereotypes encoded within the models. The prevalence of ``moving goods'' in Africa, for instance, may indicate the models' tendency to associate this region with informal commerce, warranting careful examination of whether these predictions align with ground-truth annotations or represent model biases.
To further validate this finding, we inspect different model outputs in the appendix. 

\section{Conclusion}

In this work, we present \dataset{}, a dataset capturing social group behavior from first-person view walking-tour videos across diverse urban environments worldwide. We introduce a benchmark for social group detection that poses significant challenges to existing methods, enabling fine-grained performance analysis across crowd density levels and cultural regions, two aspects largely overlooked in prior work. To establish baseline performance, we leverage state-of-the-art VLMs and LLMs, outperforming existing baselines in both AP and F1 metrics, demonstrating the effectiveness of VLM/LLMs. Beyond performance, our analysis reveals meaningful cultural variations in activity patterns and social interactions across geographic regions, opening promising directions for studying cross-cultural embodied social behavior in vision and language models.\\

\noindent
\textbf{Acknowledgments.}\\
This material is based upon work supported by the SUNY AI Platform on Google Cloud (Phases 1\&2). 

\newpage

\section{Ethics Statement}

\textbf{Data Collection and Privacy.}
Our dataset consists of publicly available YouTube walking-tour videos. We do not attempt to identify individuals in the videos, and our annotations focus solely on group-level behaviors rather than individual identification. Furthermore, we make sure that all videos were filmed only in public spaces. \dataset{} is intended solely for the study of social groups and activity recognition within these social groups. We do not provide annotations for facial recognition or re-identification. We acknowledge that group detection could potentially be misused for surveillance or discriminatory purposes. We explicitly discourage the use of our dataset for such applications, and we emphasize \dataset{} for assistive technology and human-computer interaction rather than monitoring social groups in crowded spaces.

\noindent
\textbf{Bias and Fairness.} Our analysis of cultural patterns in social behaviors is intended to uncover whether model predictions may reflect training data biases rather than genuine cultural patterns. The observed activity patterns across regions (Section 6) represent correlations in our specific dataset and should not be interpreted as definitive cultural characteristics. 

\noindent
\textbf{Annotations.} Our dataset labeling was conducted by university volunteers who were informed about the research purpose and were provided appropriate training.

\bibliographystyle{splncs04}
\bibliography{main}

@inproceedings{JRDB-Act,
    title={JRDB-Act: A Large-Scale Dataset for Spatio-Temporal Action, Social Group and Activity Detection},
    author={Ehsanpour, Mahsa and Saleh, Fatemeh and Savarese, Silvio and Reid, Ian and Rezatofighi, Hamid},
    booktitle={Proceedings of the IEEE/CVF Conference on Computer Vision and Pattern Recognition},
    year={2022}
}

@inproceedings{eccv_groups,
author = {Ehsanpour, Mahsa and Abedin, Alireza and Saleh, Fatemeh and Shi, Javen and Reid, Ian and Rezatofighi, Hamid},
title = {Joint Learning of Social Groups, Individuals Action and Sub-group Activities in Videos},
year = {2020},
isbn = {978-3-030-58544-0},
publisher = {Springer-Verlag},
address = {Berlin, Heidelberg},
url = {https://doi.org/10.1007/978-3-030-58545-7_11},
doi = {10.1007/978-3-030-58545-7_11},
booktitle = {Computer Vision – ECCV 2020: 16th European Conference, Glasgow, UK, August 23–28, 2020, Proceedings, Part IX},
pages = {177–195},
numpages = {19},
location = {Glasgow, United Kingdom}
}

@misc{piccinelli2025unidepthv2,
      title={{U}ni{D}epth{V2}: Universal Monocular Metric Depth Estimation Made Simpler}, 
      author={Luigi Piccinelli and Christos Sakaridis and Yung-Hsu Yang and Mattia Segu and Siyuan Li and Wim Abbeloos and Luc Van Gool},
      year={2025},
      eprint={2502.20110},
      archivePrefix={arXiv},
      primaryClass={cs.CV},
      url={https://arxiv.org/abs/2502.20110}, 
}

@article{zhang2025detect,
  title={Detect Anything 3D in the Wild},
  author={Zhang, Hanxue and Jiang, Haoran and Yao, Qingsong and Sun, Yanan and Zhang, Renrui and Zhao, Hao and Li, Hongyang and Zhu, Hongzi and Yang, Zetong},
  journal={arXiv preprint arXiv:2504.07958},
  year={2025}
}

@article{li2025sekai,
      title={Sekai: A Video Dataset towards World Exploration}, 
      author={Zhen Li and Chuanhao Li and Xiaofeng Mao and Shaoheng Lin and Ming Li and Shitian Zhao and Zhaopan Xu and Xinyue Li and Yukang Feng and Jianwen Sun and Zizhen Li and Fanrui Zhang and Jiaxin Ai and Zhixiang Wang and Yuwei Wu and Tong He and Jiangmiao Pang and Yu Qiao and Yunde Jia and Kaipeng Zhang},
      journal={arXiv preprint arXiv:2506.15675},
      year={2025}
}

@INPROCEEDINGS{ego4D,
  author={Grauman, Kristen and Westbury, Andrew and Byrne, Eugene and Chavis, Zachary and Furnari, Antonino and Girdhar, Rohit and Hamburger, Jackson and Jiang, Hao and Liu, Miao and Liu, Xingyu and Martin, Miguel and Nagarajan, Tushar and Radosavovic, Ilija and Ramakrishnan, Santhosh Kumar and Ryan, Fiona and Sharma, Jayant and Wray, Michael and Xu, Mengmeng and Xu, Eric Zhongcong and Zhao, Chen and Bansal, Siddhant and Batra, Dhruv and Cartillier, Vincent and Crane, Sean and Do, Tien and Doulaty, Morrie and Erapalli, Akshay and Feichtenhofer, Christoph and Fragomeni, Adriano and Fu, Qichen and Gebreselasie, Abrham and González, Cristina and Hillis, James and Huang, Xuhua and Huang, Yifei and Jia, Wenqi and Khoo, Weslie and Koláĭ, Jáchym and Kottur, Satwik and Kumar, Anurag and Landini, Federico and Li, Chao and Li, Yanghao and Li, Zhenqiang and Mangalam, Karttikeya and Modhugu, Raghava and Munro, Jonathan and Murrell, Tullie and Nishiyasu, Takumi and Price, Will and Puentes, Paola Ruiz and Ramazanova, Merey and Sari, Leda and Somasundaram, Kiran and Southerland, Audrey and Sugano, Yusuke and Tao, Ruijie and Vo, Minh and Wang, Yuchen and Wu, Xindi and Yagi, Takuma and Zhao, Ziwei and Zhu, Yunyi and Arbeláez, Pablo and Crandall, David and Damen, Dima and Farinella, Giovanni Maria and Fuegen, Christian and Ghanem, Bernard and Ithapu, Vamsi Krishna and Jawahar, C. V. and Joo, Hanbyul and Kitani, Kris and Li, Haizhou and Newcombe, Richard and Oliva, Aude and Park, Hyun Soo and Rehg, James M. and Sato, Yoichi and Shi, Jianbo and Shou, Mike Zheng and Torralba, Antonio and Torresani, Lorenzo and Yan, Mingfei and Malik, Jitendra},
  booktitle={2022 IEEE/CVF Conference on Computer Vision and Pattern Recognition (CVPR)}, 
  title={Ego4D: Around the World in 3,000 Hours of Egocentric Video}, 
  year={2022},
  volume={},
  number={},
  pages={18973-18990},
  doi={10.1109/CVPR52688.2022.01842}}

@ARTICLE{salsa_dataset,
author={Alameda-Pineda, Xavier and Staiano, Jacopo and Subramanian, Ramanathan and Batrinca, Ligia and Ricci, Elisa and Lepri, Bruno and Lanz, Oswald and Sebe, Nicu},
journal={ IEEE Transactions on Pattern Analysis \& Machine Intelligence },
title={{ SALSA: A Novel Dataset for Multimodal Group Behavior Analysis }},
year={2016},
volume={38},
number={08},
ISSN={1939-3539},
pages={1707-1720},
doi={10.1109/TPAMI.2015.2496269},
url = {https://doi.ieeecomputersociety.org/10.1109/TPAMI.2015.2496269},
publisher={IEEE Computer Society},
address={Los Alamitos, CA, USA},
month=aug}

@article{beaulieu2004intercultural,
  title={Intercultural study of personal space: A case study},
  author={Beaulieu, Catherine},
  journal={Journal of applied social psychology},
  volume={34},
  number={4},
  pages={794--805},
  year={2004},
  publisher={Wiley Online Library}
}

@article{lomranz1976cultural,
  title={Cultural variations in personal space},
  author={Lomranz, Jacob},
  journal={The Journal of Social Psychology},
  volume={99},
  number={1},
  pages={21--27},
  year={1976},
  publisher={Taylor \& Francis}
}

@incollection{aiello1980personal,
  title={Personal space, crowding, and spatial behavior in a cultural context},
  author={Aiello, John R and Thompson, Donna E},
  booktitle={Environment and culture},
  pages={107--178},
  year={1980},
  publisher={Springer}
}

@article{evans1973personal,
  title={Personal space.},
  author={Evans, Gary W and Howard, Roger B},
  journal={Psychological bulletin},
  volume={80},
  number={4},
  pages={334},
  year={1973},
  publisher={American Psychological Association}
}

@inproceedings{ge2009automatically,
  title={Automatically detecting the small group structure of a crowd},
  author={Ge, Weina and Collins, Robert T and Ruback, Barry},
  booktitle={2009 workshop on applications of computer vision (WACV)},
  pages={1--8},
  year={2009},
  organization={IEEE}
}

@inproceedings{khan2015detection,
  title={Detection of social groups in pedestrian crowds using computer vision},
  author={Khan, Sultan Daud and Vizzari, Giuseppe and Bandini, Stefania and Basalamah, Saleh},
  booktitle={Advanced Concepts for Intelligent Vision Systems: 16th International Conference, ACIVS 2015, Catania, Italy, October 26-29, 2015. Proceedings 16},
  pages={249--260},
  year={2015},
  organization={Springer}
}

@inproceedings{sandikci2011detection,
  title={Detection of human groups in videos},
  author={Sand{\i}kc{\i}, Sel{\c{c}}uk and Zinger, Svitlana and de With, Peter HN},
  booktitle={Advanced Concepts for Intelligent Vision Systems: 13th International Conference, ACIVS 2011, Ghent, Belgium, August 22-25, 2011. Proceedings 13},
  pages={507--518},
  year={2011},
  organization={Springer}
}

@inproceedings{solera2013structured,
  title={Structured learning for detection of social groups in crowd},
  author={Solera, Francesco and Calderara, Simone and Cucchiara, Rita},
  booktitle={2013 10th IEEE international conference on advanced video and signal based surveillance},
  pages={7--12},
  year={2013},
  organization={IEEE}
}

@inproceedings{ramanathan2013social,
  title={Social role discovery in human events},
  author={Ramanathan, Vignesh and Yao, Bangpeng and Fei-Fei, Li},
  booktitle={Proceedings of the IEEE conference on computer vision and pattern recognition},
  pages={2475--2482},
  year={2013}
}

@inproceedings{yu2009monitoring,
  title={Monitoring, recognizing and discovering social networks},
  author={Yu, Ting and Lim, Ser-Nam and Patwardhan, Kedar and Krahnstoever, Nils},
  booktitle={2009 IEEE Conference on Computer Vision and Pattern Recognition},
  pages={1462--1469},
  year={2009},
  organization={IEEE}
}

@inproceedings{zaidenberg2012generic,
  title={A generic framework for video understanding applied to group behavior recognition},
  author={Zaidenberg, Sofia and Boulay, Bernard and Br{\'e}mond, Fran{\c{c}}ois},
  booktitle={2012 IEEE Ninth International Conference on Advanced Video and Signal-Based Surveillance},
  pages={136--142},
  year={2012},
  organization={IEEE}
}

@inproceedings{inaba2016conversational,
  title={Conversational group detection based on social context using graph clustering algorithm},
  author={Inaba, Shoichi and Aoki, Yoshimitsu},
  booktitle={2016 12th International Conference on Signal-Image Technology \& Internet-Based Systems (SITIS)},
  pages={526--531},
  year={2016},
  organization={IEEE}
}

@inproceedings{thompson2021conversational,
  title={Conversational group detection with graph neural networks},
  author={Thompson, Sydney and Gupta, Abhijit and Gupta, Anjali W and Chen, Austin and V{\'a}zquez, Marynel},
  booktitle={Proceedings of the 2021 International Conference on Multimodal Interaction},
  pages={248--252},
  year={2021}
}

@inproceedings{alahi2016social,
  title={Social lstm: Human trajectory prediction in crowded spaces},
  author={Alahi, Alexandre and Goel, Kratarth and Ramanathan, Vignesh and Robicquet, Alexandre and Fei-Fei, Li and Savarese, Silvio},
  booktitle={Proceedings of the IEEE conference on computer vision and pattern recognition},
  pages={961--971},
  year={2016}
}

@inproceedings{alahi2014socially,
  title={Socially-aware large-scale crowd forecasting},
  author={Alahi, Alexandre and Ramanathan, Vignesh and Fei-Fei, Li},
  booktitle={Proceedings of the IEEE Conference on Computer Vision and Pattern Recognition},
  pages={2203--2210},
  year={2014}
}

@inproceedings{bisagno2018group,
  title={Group lstm: Group trajectory prediction in crowded scenarios},
  author={Bisagno, Niccol{\'o} and Zhang, Bo and Conci, Nicola},
  booktitle={Proceedings of the European conference on computer vision (ECCV) workshops},
  pages={0--0},
  year={2018}
}

@inproceedings{liang2019peeking,
  title={Peeking into the future: Predicting future person activities and locations in videos},
  author={Liang, Junwei and Jiang, Lu and Niebles, Juan Carlos and Hauptmann, Alexander G and Fei-Fei, Li},
  booktitle={Proceedings of the IEEE/CVF conference on computer vision and pattern recognition},
  pages={5725--5734},
  year={2019}
}

@inproceedings{liang2020garden,
  title={The garden of forking paths: Towards multi-future trajectory prediction},
  author={Liang, Junwei and Jiang, Lu and Murphy, Kevin and Yu, Ting and Hauptmann, Alexander},
  booktitle={Proceedings of the IEEE/CVF Conference on Computer Vision and Pattern Recognition},
  pages={10508--10518},
  year={2020}
}

@inproceedings{hu2020probabilistic,
  title={Probabilistic future prediction for video scene understanding},
  author={Hu, Anthony and Cotter, Fergal and Mohan, Nikhil and Gurau, Corina and Kendall, Alex},
  booktitle={Computer Vision--ECCV 2020: 16th European Conference, Glasgow, UK, August 23--28, 2020, Proceedings, Part XVI 16},
  pages={767--785},
  year={2020},
  organization={Springer}
}

@article{hassan2024predicting,
  title={Predicting humans future motion trajectories in video streams using generative adversarial network},
  author={Hassan, Muhammad Ahmed and Khan, Muhammad Usman Ghani and Iqbal, Razi and Riaz, Omer and Bashir, Ali Kashif and Tariq, Usman},
  journal={Multimedia Tools and Applications},
  volume={83},
  number={5},
  pages={15289--15311},
  year={2024},
  publisher={Springer}
}

@article{afsar2018automatic,
  title={Automatic human trajectory destination prediction from video},
  author={Afsar, Palwasha and Cortez, Paulo and Santos, Henrique},
  journal={Expert Systems with Applications},
  volume={110},
  pages={41--51},
  year={2018},
  publisher={Elsevier}
}

@inproceedings{bi2020can,
  title={How can i see my future? fvtraj: Using first-person view for pedestrian trajectory prediction},
  author={Bi, Huikun and Zhang, Ruisi and Mao, Tianlu and Deng, Zhigang and Wang, Zhaoqi},
  booktitle={Computer Vision--ECCV 2020: 16th European Conference, Glasgow, UK, August 23--28, 2020, Proceedings, Part VII 16},
  pages={576--593},
  year={2020},
  organization={Springer}
}

@article{wu2021comprehensive,
  title={A comprehensive review of group activity recognition in videos},
  author={Wu, Li-Fang and Wang, Qi and Jian, Meng and Qiao, Yu and Zhao, Bo-Xuan},
  journal={International Journal of Automation and Computing},
  volume={18},
  number={3},
  pages={334--350},
  year={2021},
  publisher={Springer}
}

@inproceedings{bazzani2012decentralized,
  title={Decentralized particle filter for joint individual-group tracking},
  author={Bazzani, Loris and Cristani, Marco and Murino, Vittorio},
  booktitle={2012 IEEE Conference on Computer Vision and Pattern Recognition},
  pages={1886--1893},
  year={2012},
  organization={IEEE}
}

@article{bazzani2014joint,
  title={Joint individual-group modeling for tracking},
  author={Bazzani, Loris and Zanotto, Matteo and Cristani, Marco and Murino, Vittorio},
  journal={IEEE transactions on pattern analysis and machine intelligence},
  volume={37},
  number={4},
  pages={746--759},
  year={2014},
  publisher={IEEE}
}

@article{veeraraghavan2003computer,
  title={Computer vision algorithms for intersection monitoring},
  author={Veeraraghavan, Harini and Masoud, Osama and Papanikolopoulos, Nikolaos P},
  journal={IEEE Transactions on Intelligent Transportation Systems},
  volume={4},
  number={2},
  pages={78--89},
  year={2003},
  publisher={IEEE}
}

@article{brunetti2018computer,
  title={Computer vision and deep learning techniques for pedestrian detection and tracking: A survey},
  author={Brunetti, Antonio and Buongiorno, Domenico and Trotta, Gianpaolo Francesco and Bevilacqua, Vitoantonio},
  journal={Neurocomputing},
  volume={300},
  pages={17--33},
  year={2018},
  publisher={Elsevier}
}

@article{globe,
title = {GLOBE: A twenty year journey into the intriguing world of culture and leadership},
journal = {Journal of World Business},
volume = {47},
number = {4},
pages = {504-518},
year = {2012},
note = {SPECIAL ISSUE: LEADERSHIP IN A GLOBAL CONTEXT},
issn = {1090-9516},
doi = {https://doi.org/10.1016/j.jwb.2012.01.004},
url = {https://www.sciencedirect.com/science/article/pii/S1090951612000053},
author = {Peter Dorfman and Mansour Javidan and Paul Hanges and Ali Dastmalchian and Robert House}
}

@misc{liu2025minglevlmssemanticallycomplex,
      title={MINGLE: VLMs for Semantically Complex Region Detection in Urban Scenes}, 
      author={Liu Liu and Alexandra Kudaeva and Marco Cipriano and Fatimeh Al Ghannam and Freya Tan and Gerard de Melo and Andres Sevtsuk},
      year={2025},
      eprint={2509.13484},
      archivePrefix={arXiv},
      primaryClass={cs.CV},
      url={https://arxiv.org/abs/2509.13484}, 
}

@INPROCEEDINGS{PANDA,
  author={Wang, Xueyang and Zhang, Xiya and Zhu, Yinheng and Guo, Yuchen and Yuan, Xiaoyun and Xiang, Liuyu and Wang, Zerun and Ding, Guiguang and Brady, David and Dai, Qionghai and Fang, Lu},
  booktitle={2020 IEEE/CVF Conference on Computer Vision and Pattern Recognition (CVPR)}, 
  title={PANDA: A Gigapixel-Level Human-Centric Video Dataset}, 
  year={2020},
  volume={},
  number={},
  pages={3265-3275},
  doi={10.1109/CVPR42600.2020.00333}}

@article{SCU-VSD,
title = {A new approach for social group detection based on spatio-temporal interpersonal distance measurement},
journal = {Heliyon},
volume = {8},
number = {10},
pages = {e11038},
year = {2022},
issn = {2405-8440},
doi = {https://doi.org/10.1016/j.heliyon.2022.e11038},
url = {https://www.sciencedirect.com/science/article/pii/S240584402202326X},
author = {Jie Su and Jianglan Huang and Linbo Qing and Xiaohai He and Honggang Chen}
}

@ARTICLE{GVEII,
  author={Solera, Francesco and Calderara, Simone and Cucchiara, Rita},
  journal={IEEE Transactions on Pattern Analysis and Machine Intelligence}, 
  title={Socially Constrained Structural Learning for Groups Detection in Crowd}, 
  year={2016},
  volume={38},
  number={5},
  pages={995-1008},
  doi={10.1109/TPAMI.2015.2470658}}

@INPROCEEDINGS{CUB,
  author={Wongun Choi and Shahid, Khuram and Savarese, Silvio},
  booktitle={2009 IEEE 12th International Conference on Computer Vision Workshops, ICCV Workshops}, 
  title={What are they doing? : Collective activity classification using spatio-temporal relationship among people}, 
  year={2009},
  volume={},
  number={},
  pages={1282-1289},
  doi={10.1109/ICCVW.2009.5457461}}

@inproceedings{khattab2024dspy,
  title={DSPy: Compiling Declarative Language Model Calls into Self-Improving Pipelines},
  author={Khattab, Omar and Singhvi, Arnav and Maheshwari, Paridhi and Zhang, Zhiyuan and Santhanam, Keshav and Vardhamanan, Sri and Haq, Saiful and Sharma, Ashutosh and Joshi, Thomas T. and Moazam, Hanna and Miller, Heather and Zaharia, Matei and Potts, Christopher},
  journal={The Twelfth International Conference on Learning Representations},
  year={2024}
}

@inproceedings{liu2025learningpedgen,
title={Learning to Generate Diverse Pedestrian Movements from Web Videos with Noisy Labels},
author={Liu, Zhizheng and Lin, Joe and Wu, Wayne and Zhou, Bolei},
booktitle={The Thirteenth International Conference on Learning Representations},
year={2025}
}

@inproceedings{bae2022gpgraph,
  title={Learning Pedestrian Group Representations for Multi-modal Trajectory Prediction},
  author={Bae, Inhwan and Park, Jin-Hwi and Jeon, Hae-Gon},
  booktitle={Proceedings of the European Conference on Computer Vision},
  year={2022}
}

@article{kothari2022traj,
author = {Kothari, Parth and Kreiss, Sven and Alahi, Alexandre},
title = {Human Trajectory Forecasting in Crowds: A Deep Learning Perspective},
year = {2022},
issue_date = {July 2022},
publisher = {IEEE Press},
volume = {23},
number = {7},
issn = {1524-9050},
url = {https://doi.org/10.1109/TITS.2021.3069362},
doi = {10.1109/TITS.2021.3069362},
journal = {Trans. Intell. Transport. Sys.},
month = jul,
pages = {7386–7400},
numpages = {15}
}

@INPROCEEDINGS{shafiee2021introvert,
  author={Shafiee, Nasim and Padir, Taskin and Elhamifar, Ehsan},
  booktitle={2021 IEEE/CVF Conference on Computer Vision and Pattern Recognition (CVPR)}, 
  title={Introvert: Human Trajectory Prediction via Conditional 3D Attention}, 
  year={2021},
  volume={},
  number={},
  pages={16810-16820},
  doi={10.1109/CVPR46437.2021.01654}}

@inproceedings{gupta2018social,
  title={Social GAN: Socially Acceptable Trajectories with Generative Adversarial Networks},
  author={Gupta, Agrim and Johnson, Justin and Fei-Fei, Li and Savarese, Silvio and Alahi, Alexandre},
  booktitle={IEEE Conference on Computer Vision and Pattern Recognition (CVPR)},
  number={CONF},
  year={2018}
}

@inproceedings{VHELM,
author = {Lee, Tony and Tu, Haoqin and Wong, Chi Heem and Zheng, Wenhao and Zhou, Yiyang and Mai, Yifan and Roberts, Josselin Somerville and Yasunaga, Michihiro and Yao, Huaxiu and Xie, Cihang and Liang, Percy},
title = {VHELM: a holistic evaluation of vision language models},
year = {2024},
isbn = {9798331314385},
publisher = {Curran Associates Inc.},
address = {Red Hook, NY, USA},
booktitle = {Proceedings of the 38th International Conference on Neural Information Processing Systems},
articleno = {4464},
numpages = {35},
location = {Vancouver, BC, Canada},
series = {NIPS '24}
}

@inproceedings{VISBIAS,
    title = "{V}is{B}ias: Measuring Explicit and Implicit Social Biases in Vision Language Models",
    author = "Huang, Jen-tse  and
      Qin, Jiantong  and
      Zhang, Jianping  and
      Yuan, Youliang  and
      Wang, Wenxuan  and
      Zhao, Jieyu",
    editor = "Christodoulopoulos, Christos  and
      Chakraborty, Tanmoy  and
      Rose, Carolyn  and
      Peng, Violet",
    booktitle = "Proceedings of the 2025 Conference on Empirical Methods in Natural Language Processing",
    month = nov,
    year = "2025",
    address = "Suzhou, China",
    publisher = "Association for Computational Linguistics",
    url = "https://aclanthology.org/2025.emnlp-main.908/",
    doi = "10.18653/v1/2025.emnlp-main.908",
    pages = "17981--18004",
    ISBN = "979-8-89176-332-6"
}

@inproceedings{culturalVQA,
    title = "Benchmarking Vision Language Models for Cultural Understanding",
    author = "Nayak, Shravan  and
      Jain, Kanishk  and
      Awal, Rabiul  and
      Reddy, Siva  and
      Steenkiste, Sjoerd Van  and
      Hendricks, Lisa Anne  and
      Stanczak, Karolina  and
      Agrawal, Aishwarya",
    editor = "Al-Onaizan, Yaser  and
      Bansal, Mohit  and
      Chen, Yun-Nung",
    booktitle = "Proceedings of the 2024 Conference on Empirical Methods in Natural Language Processing",
    month = nov,
    year = "2024",
    address = "Miami, Florida, USA",
    publisher = "Association for Computational Linguistics",
    url = "https://aclanthology.org/2024.emnlp-main.329/",
    doi = "10.18653/v1/2024.emnlp-main.329",
    pages = "5769--5790"
}

@inproceedings{CulturalBench,
    title = "{C}ultural{B}ench: A Robust, Diverse and Challenging Benchmark for Measuring {LM}s' Cultural Knowledge Through Human-{AI} Red-Teaming",
    author = "Chiu, Yu Ying  and
      Jiang, Liwei  and
      Lin, Bill Yuchen  and
      Park, Chan Young  and
      Li, Shuyue Stella  and
      Ravi, Sahithya  and
      Bhatia, Mehar  and
      Antoniak, Maria  and
      Tsvetkov, Yulia  and
      Shwartz, Vered  and
      Choi, Yejin",
    editor = "Che, Wanxiang  and
      Nabende, Joyce  and
      Shutova, Ekaterina  and
      Pilehvar, Mohammad Taher",
    booktitle = "Proceedings of the 63rd Annual Meeting of the Association for Computational Linguistics (Volume 1: Long Papers)",
    month = jul,
    year = "2025",
    address = "Vienna, Austria",
    publisher = "Association for Computational Linguistics",
    url = "https://aclanthology.org/2025.acl-long.1247/",
    doi = "10.18653/v1/2025.acl-long.1247",
    pages = "25663--25701",
    ISBN = "979-8-89176-251-0"
}

@misc{blendvisbenchmarkingmultimodalcultural,
      title={BLEnD-Vis: Benchmarking Multimodal Cultural Understanding in Vision Language Models}, 
      author={Bryan Chen Zhengyu Tan and Zheng Weihua and Zhengyuan Liu and Nancy F. Chen and Hwaran Lee and Kenny Tsu Wei Choo and Roy Ka-Wei Lee},
      year={2025},
      eprint={2510.11178},
      archivePrefix={arXiv},
      primaryClass={cs.CV},
      url={https://arxiv.org/abs/2510.11178}, 
}

@INPROCEEDINGS{all_languages_matter,

  author={Vayani, Ashmal and Dissanayake, Dinura and Watawana, Hasindri and Ahsan, Noor and Sasikumar, Nevasini and Thawakar, Omkar and Ademtew, Henok Biadglign and Hmaiti, Yahya and Kumar, Amandeep and Kuckreja, Kartik and Maslych, Mykola and Al Ghallabi, Wafa and Mihaylov, Mihail and Qin, Chao and Shaker, Abdelrahman M and Zhang, Mike and Ihsani, Mahardika Krisna and Esplana, Amiel and Gokani, Monil and Mirkin, Shachar and Singh, Harsh and Srivastava, Ashay and Hamerlik, Endre and Asma Izzati, Fathinah and Adamsyah Maani, Fadillah and Cavada, Sebastian and Chim, Jenny and Gupta, Rohit and Manjunath, Sanjay and Zhumakhanova, Kamila and Heriniaina Rabevohitra, Feno and Amirudin, Azril and Ridzuan, Muhammad and Kareem, Daniya and More, Ketan and Li, Kunyang and Shakya, Pramesh and Saad, Muhammad and Ghasemaghaei, Amirpouya and Djanibekov, Amirbek and Azizov, Dilshod and Jankovic, Branislava and Bhatia, Naman and Cabrera, Alvaro and Obando-Ceron, Johan and Otieno, Olympiah and Farestam, Fabian and Rabbani, Muztoba and Baliah, Sanoojan and Sanjeev, Santosh and Shtanchaev, Abduragim and Fatima, Maheen and Nguyen, Thao and Kareem, Amrin and Aremu, Toluwani and Xavier, Nathan and Bhatkal, Amit and Toyin, Hawau and Chadha, Aman and Cholakkal, Hisham and Anwer, Rao Muhammad and Felsberg, Michael and Laaksonen, Jorma and Solorio, Thamar and Choudhury, Monojit and Laptev, Ivan and Shah, Mubarak and Khan, Salman and Khan, Fahad Shahbaz},

  booktitle={2025 IEEE/CVF Conference on Computer Vision and Pattern Recognition (CVPR)}, 

  title={All Languages Matter: Evaluating LMMs on Culturally Diverse 100 Languages}, 

  year={2025},

  volume={},

  number={},

  pages={19565-19575},

  doi={10.1109/CVPR52734.2025.01822}}

@inproceedings{CANDLE,
author = {Nguyen, Tuan-Phong and Razniewski, Simon and Varde, Aparna and Weikum, Gerhard},
title = {Extracting Cultural Commonsense Knowledge at Scale},
year = {2023},
isbn = {9781450394161},
publisher = {Association for Computing Machinery},
address = {New York, NY, USA},
url = {https://doi.org/10.1145/3543507.3583535},
doi = {10.1145/3543507.3583535},
booktitle = {Proceedings of the ACM Web Conference 2023},
pages = {1907–1917},
numpages = {11},
location = {Austin, TX, USA},
series = {WWW '23}
}

@inproceedings{CVQA,
author = {Romero, David and Lyu, Chenyang and Wibowo, Haryo Akbarianto and Lynn, Teresa and Hamed, Injy and Kishore, Aditya Nanda and Mandal, Aishik and Dragonetti, Alina and Abzaliev, Artem and Tonja, Atnafu Lambebo and Balcha, Bontu Fufa and Whitehouse, Chenxi and Salamea, Christian and Velasco, Dan John and Adelani, David Ifeoluwa and Le Meur, David and Villa-Cueva, Emilio and Koto, Fajri and Farooqui, Fauzan and Belcavello, Frederico and Batnasan, Ganzorig and Vallejo, Gisela and Caulfield, Grainne and Ivetta, Guido and Song, Haiyue and Ademtew, Henok Biadglign and Maina, Hern\'{a}n and Lovenia, Holy and Azime, Israel Abebe and Cruz, Jan Christian Blaise and Gala, Jay and Geng, Jiahui and Ortiz-Barajas, Jesus-German and Baek, Jinheon and Dunstan, Jocelyn and Alemany, Laura Alonso and Nagasinghe, Kumaranage Ravindu Yasas and Benotti, Luciana and D'Haro, Luis Fernando and Viridiano, Marcelo and Estecha-Garitagoitia, Marcos and Cabrera, Maria Camila Buitrago and Rodr\'{\i}guez-Cantelar, Mario and Jouitteau, M\'{e}lanie and Mihaylov, Mihail and Etori, Naome and Imam, Mohamed Fazli Mohamed and Adilazuarda, Muhammad Farid and Gochoo, Munkhjargal and Otgonbold, Munkh-Erdene and Niyomugisha, Olivier and Silva, Paula M\'{o}nica and Chitale, Pranjal and Dabre, Raj and Chevi, Rendi and Zhang, Ruochen and Diandaru, Ryandito and Cahyawijaya, Samuel and G\'{o}ngora, Santiago and Jeong, Soyeong and Purkayastha, Sukannya and Kuribayashi, Tatsuki and Clifford, Teresa and Jayakumar, Thanmay and Torrent, Tiago Timponi and Ehsan, Toqeer and Araujo, Vladimir and Kementchedjhieva, Yova and Burzo, Zara and Lim, Zheng Wei and Yong, Zheng Xin and Ignat, Oana and Nwatu, Joan and Mihalcea, Rada and Solorio, Thamar and Aji, Alham Fikri},
title = {CVQA: culturally-diverse multilingual visual question answering benchmark},
year = {2024},
isbn = {9798331314385},
publisher = {Curran Associates Inc.},
address = {Red Hook, NY, USA},
booktitle = {Proceedings of the 38th International Conference on Neural Information Processing Systems},
articleno = {366},
numpages = {27},
location = {Vancouver, BC, Canada},
series = {NIPS '24}
}

@misc{satar2025seeingculturebenchmarkvisual,
      title={Seeing Culture: A Benchmark for Visual Reasoning and Grounding}, 
      author={Burak Satar and Zhixin Ma and Patrick A. Irawan and Wilfried A. Mulyawan and Jing Jiang and Ee-Peng Lim and Chong-Wah Ngo},
      year={2025},
      eprint={2509.16517},
      archivePrefix={arXiv},
      primaryClass={cs.CV},
      url={https://arxiv.org/abs/2509.16517}, 
}

@inproceedings{bhatia-etal-2024-local,
    title = "From Local Concepts to Universals: Evaluating the Multicultural Understanding of Vision-Language Models",
    author = "Bhatia, Mehar  and
      Ravi, Sahithya  and
      Chinchure, Aditya  and
      Hwang, EunJeong  and
      Shwartz, Vered",
    editor = "Al-Onaizan, Yaser  and
      Bansal, Mohit  and
      Chen, Yun-Nung",
    booktitle = "Proceedings of the 2024 Conference on Empirical Methods in Natural Language Processing",
    month = nov,
    year = "2024",
    address = "Miami, Florida, USA",
    publisher = "Association for Computational Linguistics",
    url = "https://aclanthology.org/2024.emnlp-main.385/",
    doi = "10.18653/v1/2024.emnlp-main.385",
    pages = "6763--6782"
}

@inproceedings{jeong-etal-2025-culture,
    title = "Culture-{TRIP}: Culturally-Aware Text-to-Image Generation with Iterative Prompt Refinement",
    author = "Jeong, Suchae  and
      Choi, Inseong  and
      Yun, Youngsik  and
      Kim, Jihie",
    editor = "Chiruzzo, Luis  and
      Ritter, Alan  and
      Wang, Lu",
    booktitle = "Proceedings of the 2025 Conference of the Nations of the Americas Chapter of the Association for Computational Linguistics: Human Language Technologies (Volume 1: Long Papers)",
    month = apr,
    year = "2025",
    address = "Albuquerque, New Mexico",
    publisher = "Association for Computational Linguistics",
    url = "https://aclanthology.org/2025.naacl-long.483/",
    doi = "10.18653/v1/2025.naacl-long.483",
    pages = "9543--9573",
    ISBN = "979-8-89176-189-6"
}

@article{MensahChen2013,
  author       = {Mensah, Yaw M. and Chen, Hsiao{-}Yin},
  title        = {Global Clustering of Countries by Culture -- An Extension of the GLOBE Study},
  year         = {2013},
  month        = apr,
  note         = {Available at SSRN: \url{https://ssrn.com/abstract=2189904} or \url{http://dx.doi.org/10.2139/ssrn.2189904}},
  journal      = {SSRN Electronic Journal},
  doi          = {10.2139/ssrn.2189904}
}

@article{bai2025qwen2,
  title={Qwen2. 5-vl technical report},
  author={Bai, Shuai and Chen, Keqin and Liu, Xuejing and Wang, Jialin and Ge, Wenbin and Song, Sibo and Dang, Kai and Wang, Peng and Wang, Shijie and Tang, Jun and others},
  journal={arXiv preprint arXiv:2502.13923},
  year={2025}
}

@misc{qwen3,
      title={Qwen3 Technical Report}, 
      author={An Yang and Anfeng Li and Baosong Yang and Beichen Zhang and Binyuan Hui and Bo Zheng and Bowen Yu and Chang Gao and Chengen Huang and Chenxu Lv and Chujie Zheng and Dayiheng Liu and Fan Zhou and Fei Huang and Feng Hu and Hao Ge and Haoran Wei and Huan Lin and Jialong Tang and Jian Yang and Jianhong Tu and Jianwei Zhang and Jianxin Yang and Jiaxi Yang and Jing Zhou and Jingren Zhou and Junyang Lin and Kai Dang and Keqin Bao and Kexin Yang and Le Yu and Lianghao Deng and Mei Li and Mingfeng Xue and Mingze Li and Pei Zhang and Peng Wang and Qin Zhu and Rui Men and Ruize Gao and Shixuan Liu and Shuang Luo and Tianhao Li and Tianyi Tang and Wenbiao Yin and Xingzhang Ren and Xinyu Wang and Xinyu Zhang and Xuancheng Ren and Yang Fan and Yang Su and Yichang Zhang and Yinger Zhang and Yu Wan and Yuqiong Liu and Zekun Wang and Zeyu Cui and Zhenru Zhang and Zhipeng Zhou and Zihan Qiu},
      year={2025},
      eprint={2505.09388},
      archivePrefix={arXiv},
      primaryClass={cs.CL},
      url={https://arxiv.org/abs/2505.09388}, 
}

@ARTICLE{10286297,
  author={Zhang, Jinsong and Gu, Lingfeng and Lai, Yu-Kun and Wang, Xueyang and Li, Kun},
  journal={IEEE Transactions on Circuits and Systems for Video Technology}, 
  title={Toward Grouping in Large Scenes With Occlusion-Aware Spatio–Temporal Transformers}, 
  year={2024},
  volume={34},
  number={5},
  pages={3919-3929},
  doi={10.1109/TCSVT.2023.3324868}}

@InProceedings{Yokoyama_2025_ICCV,
    author    = {Yokoyama, Kaname and Nakatani, Chihiro and Ukita, Norimichi},
    title     = {Dynamic Group Detection using VLM-augmented Temporal Groupness Graph},
    booktitle = {Proceedings of the IEEE/CVF International Conference on Computer Vision (ICCV)},
    month     = {October},
    year      = {2025},
    pages     = {10475-10484}
}

@misc{carion2025sam3segmentconcepts,
      title={SAM 3: Segment Anything with Concepts},
      author={Nicolas Carion and Laura Gustafson and Yuan-Ting Hu and Shoubhik Debnath and Ronghang Hu and Didac Suris and Chaitanya Ryali and Kalyan Vasudev Alwala and Haitham Khedr and Andrew Huang and Jie Lei and Tengyu Ma and Baishan Guo and Arpit Kalla and Markus Marks and Joseph Greer and Meng Wang and Peize Sun and Roman Rädle and Triantafyllos Afouras and Effrosyni Mavroudi and Katherine Xu and Tsung-Han Wu and Yu Zhou and Liliane Momeni and Rishi Hazra and Shuangrui Ding and Sagar Vaze and Francois Porcher and Feng Li and Siyuan Li and Aishwarya Kamath and Ho Kei Cheng and Piotr Dollár and Nikhila Ravi and Kate Saenko and Pengchuan Zhang and Christoph Feichtenhofer},
      year={2025},
      eprint={2511.16719},
      archivePrefix={arXiv},
      primaryClass={cs.CV},
      url={https://arxiv.org/abs/2511.16719},
}

@inproceedings{kim2024towards,
  title     = {Towards More Practical Group Activity Detection: A New Benchmark and Model},
  author    = {Kim, Dongkeun and Song, Youngkil and Cho, Minsu and Kwak, Suha},
  booktitle = {European Conference on Computer Vision (ECCV)},
  pages     = {240--258},
  year      = {2024},
  organization = {Springer}
}

@misc{mcinnes2020umap,
      title={UMAP: Uniform Manifold Approximation and Projection for Dimension Reduction}, 
      author={Leland McInnes and John Healy and James Melville},
      year={2020},
      eprint={1802.03426},
      archivePrefix={arXiv},
      primaryClass={stat.ML},
      url={https://arxiv.org/abs/1802.03426}, 
}

\clearpage

\appendix

\section{Coarse-grained Annotation Details}

\begin{figure}[!ht]
    \centering
    \begin{subfigure}{0.45\linewidth}
        \centering
        \includegraphics[width=\linewidth]{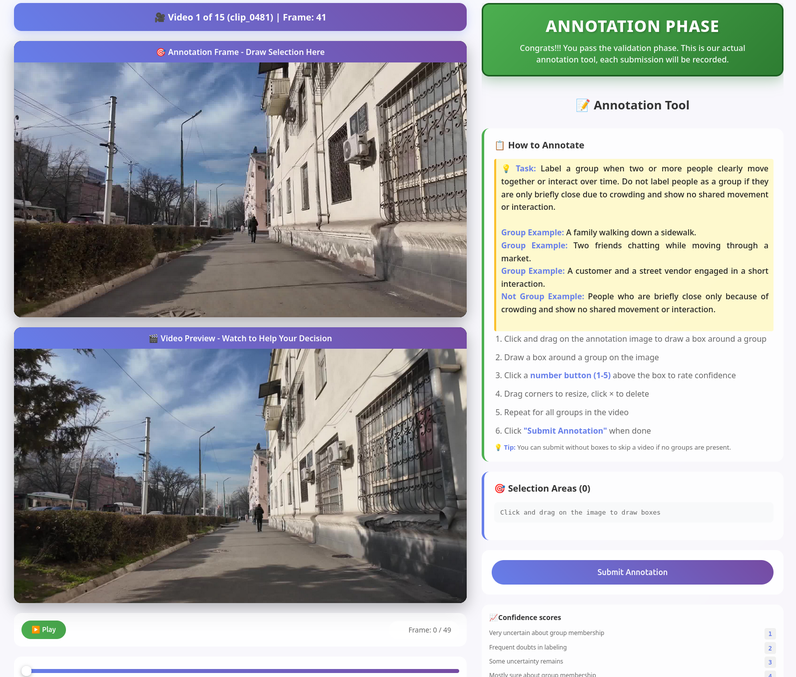}
        \caption{Coarse Annotation interface}
        \label{fig:sup_annotation_s1}
    \end{subfigure}
    \hfill
    \begin{subfigure}{0.45\linewidth}
        \centering
        \includegraphics[width=\linewidth]{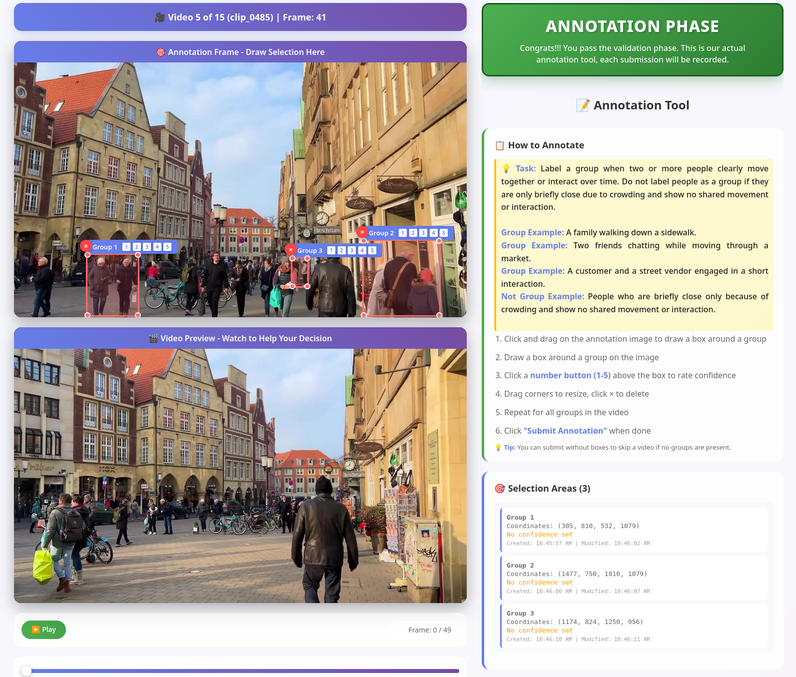}
        \caption{Coarse Annotation interface  with boxes}
        \label{fig:sup_annotation_s2}
    \end{subfigure}
    
    \caption{Coarse annotation tool used for labeling groups.}
    \label{fig:sup_annotation_tool_coarse}
\end{figure}

~\autoref{fig:sup_annotation_s1} illustrates our annotation interface for the coarse-grained annotation task. We ask annotators to create bounding boxes by clicking and dragging on the top canvas. When needed, they can review the corresponding video to verify their decisions. For each bounding box, annotators also assign a confidence score ranging from 1 to 5, following the Likert scale shown in ~\autoref{tab:sup_likert_scale}.

\begin{table}[h!]
\centering
\caption{Confidence scale for group membership labeling.}
\resizebox{12.5cm}{!}{
\begin{tabular}{cll}
\hline
\textbf{Value} & \textbf{Confidence Level} & \textbf{Description} \\ 
\hline
1 & Not confident at all     & Very uncertain about group membership \\
2 & Low confidence           & Frequent doubts in labeling \\
3 & Moderately confident     & Some uncertainty remains \\
4 & Confident                & Mostly sure about group membership \\
5 & Very confident           & Certain about group detection correctness \\
\hline
\end{tabular}
}
\label{tab:sup_likert_scale}
\end{table}

After completing a clip, annotators click “Submit annotation” to proceed to the next one, repeating this process until they finish all clips. Each submission creates a database entry that records metadata about the annotation, including the timestamp, the number of actions identified, the video viewing history, and other relevant information. For example, the median annotation time per clip is 30 seconds. In particular, annotators spend a median of 13 seconds on scattered scenes, 38 seconds on moderately populated scenes, and 54 seconds on crowded scenes.\\

For each frame, we end up with at least 3 annotators. To get a final annotation set, we iterate through each detected group and adding it only if its bounding box does not overlap with any previously included box (IoU $\geq 0.3$). We adopt this strategy to maximize group detections, as individual annotators may overlook some instances. ~\autoref{fig:coarse_finegrained_examples} (left) presents coarse annotations after this procedure.

\newpage

\section{Fine-grained Annotation Details}

\begin{figure}[!h]
    \centering
    \begin{subfigure}{0.45\linewidth}
        \centering
        \includegraphics[width=\linewidth]{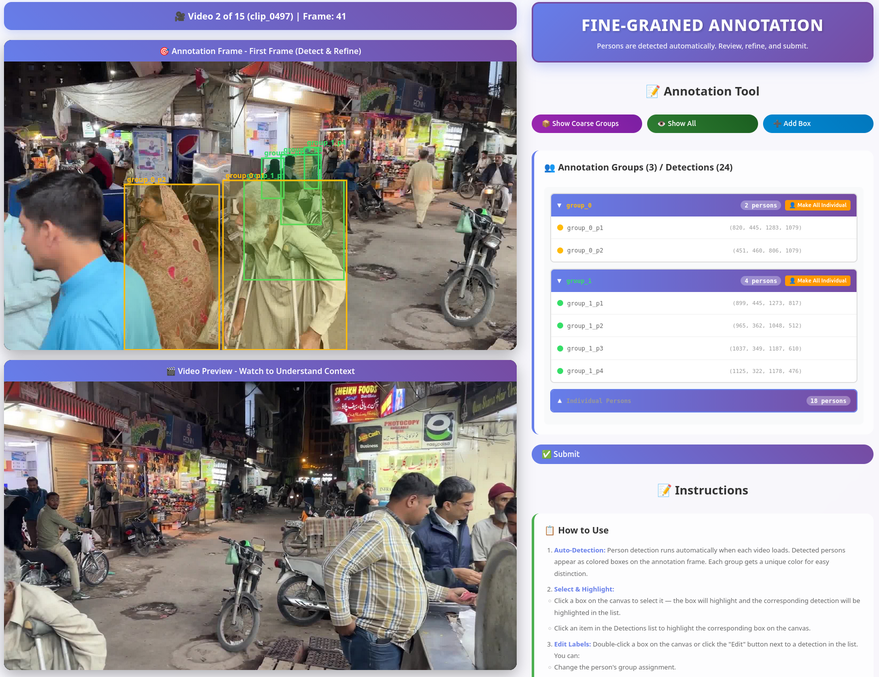}
        \caption{Fine-grained Annotation interface before refinement}
        \label{fig:sup_fine_annotation_s1}
    \end{subfigure}
    \hfill
    \begin{subfigure}{0.45\linewidth}
        \centering
        \includegraphics[width=\linewidth]{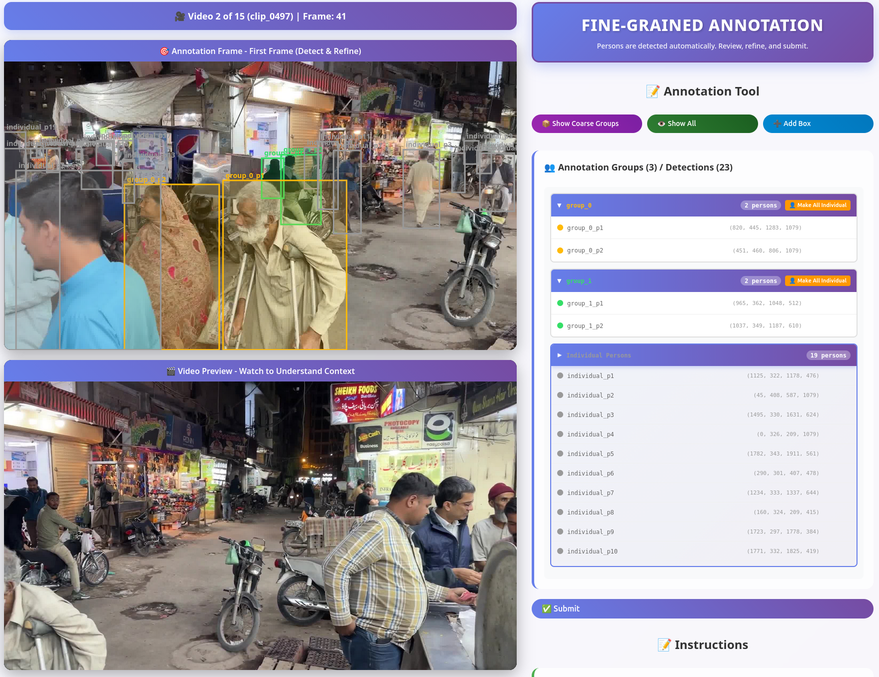}
        \caption{Fine-grained Annotation interface after refinement}
        \label{fig:sup_fine_annotation_s2}
    \end{subfigure}
    
    \caption{Fine-grained annotation tool used for labeling groups.}
    \label{fig:sup_annotation_tool_fine}
\end{figure}

\begin{figure}[!h]
    \centering
    \begin{subfigure}[b]{0.48\linewidth}
        \includegraphics[width=\linewidth]{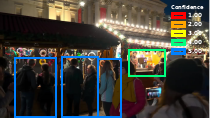}
    \end{subfigure}
    \hfill
    \begin{subfigure}[b]{0.48\linewidth}
        \includegraphics[width=\linewidth]{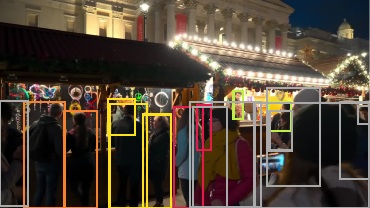}
    \end{subfigure}

    \vspace{0.5em}

    \begin{subfigure}[b]{0.48\linewidth}
        \includegraphics[width=\linewidth]{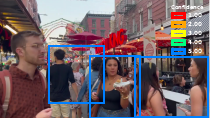}
    \end{subfigure}
    \hfill
    \begin{subfigure}[b]{0.48\linewidth}
        \includegraphics[width=\linewidth]{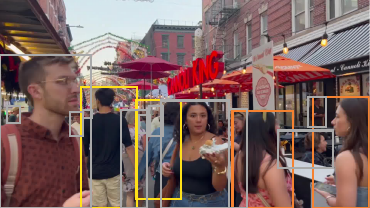}
    \end{subfigure}

    \vspace{0.5em}

    \begin{subfigure}[b]{0.48\linewidth}
        \includegraphics[width=\linewidth]{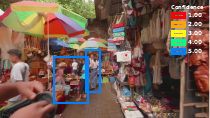}
    \end{subfigure}
    \hfill
    \begin{subfigure}[b]{0.48\linewidth}
        \includegraphics[width=\linewidth]{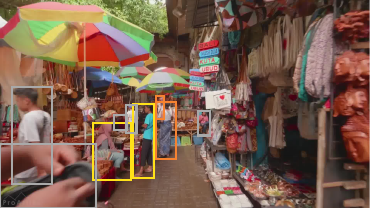}
    \end{subfigure}

    \caption{Examples of coarse (left) and fine-grained (right) annotations.}
    \label{fig:coarse_finegrained_examples}
\end{figure}

~\autoref{fig:sup_fine_annotation_s1} illustrates our interface for the fine-grained annotation task. We first propose an initial grouping based on the final set of coarse groups, and then ask annotators to refine these annotations. The initial set of bounding boxes is detected by SAM3~\cite{carion2025sam3segmentconcepts} over the full image, as well as over each coarse patch and an additional third of its surrounding area. Annotators can remove bounding boxes, modify the group label assigned to a bounding box, or add new bounding boxes. When necessary, they can review the corresponding video to support their decisions. ~\autoref{fig:sup_fine_annotation_s2} presents an example after refinement, together with the detected individuals. Similar to the coarse annotation task, clicking the submit button records the annotation in the database and loads the next clip. ~\autoref{fig:coarse_finegrained_examples} (right) shows finegrained annotations after this procedure.\\

On average, we reach three annotators for each frame, achieving an inter-annotator agreement of 95.56\% with the preceding coarse annotations. To consolidate the annotations, we apply a majority-vote scheme combined with a connected-components heuristic. Specifically, each group label must appear in at least half of the annotators’ responses, and we match candidate bounding boxes across annotators using an area-adaptive IoU criterion. We retain only connected components with at least two members and merge them to produce the final fine-grained detections, resulting in a fine-grained inter-annotator agreement of 91.64\%. Overall, the median annotation time per clip is 22 seconds. For scattered clips, the median time is 6 seconds, for moderately crowded clips 23 seconds, and for crowded clips 53 seconds.

\newpage

\section{Analysis of Social Activities}

\begin{figure}[h!] 
    \centering
    \includegraphics[width=0.93\linewidth]{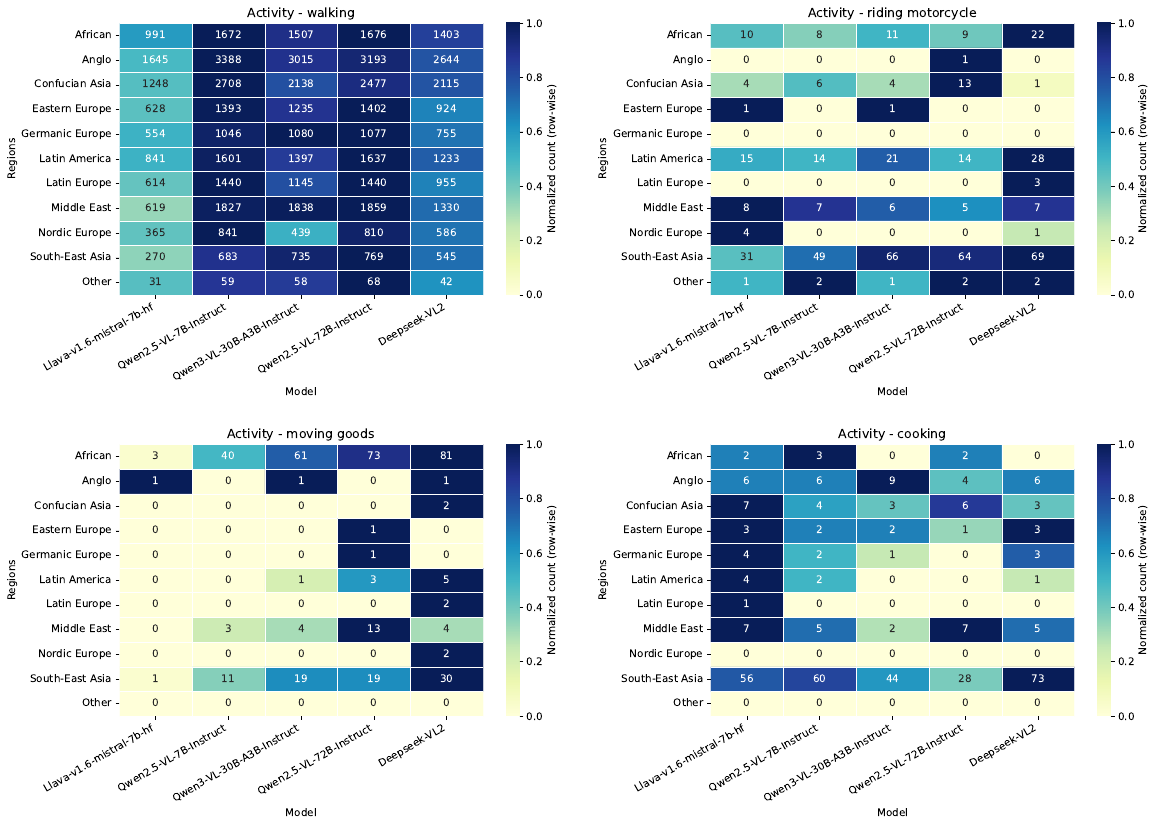}
    \caption{Regional distribution of activity recognition across VLMs for ``riding motorcycle'', ``moving goods'', and ``cooking'' .}
    \label{fig:region_vs_model_4activities_similar}
\end{figure}

\begin{figure}[t!] 
    \centering
    \includegraphics[width=0.93\linewidth]{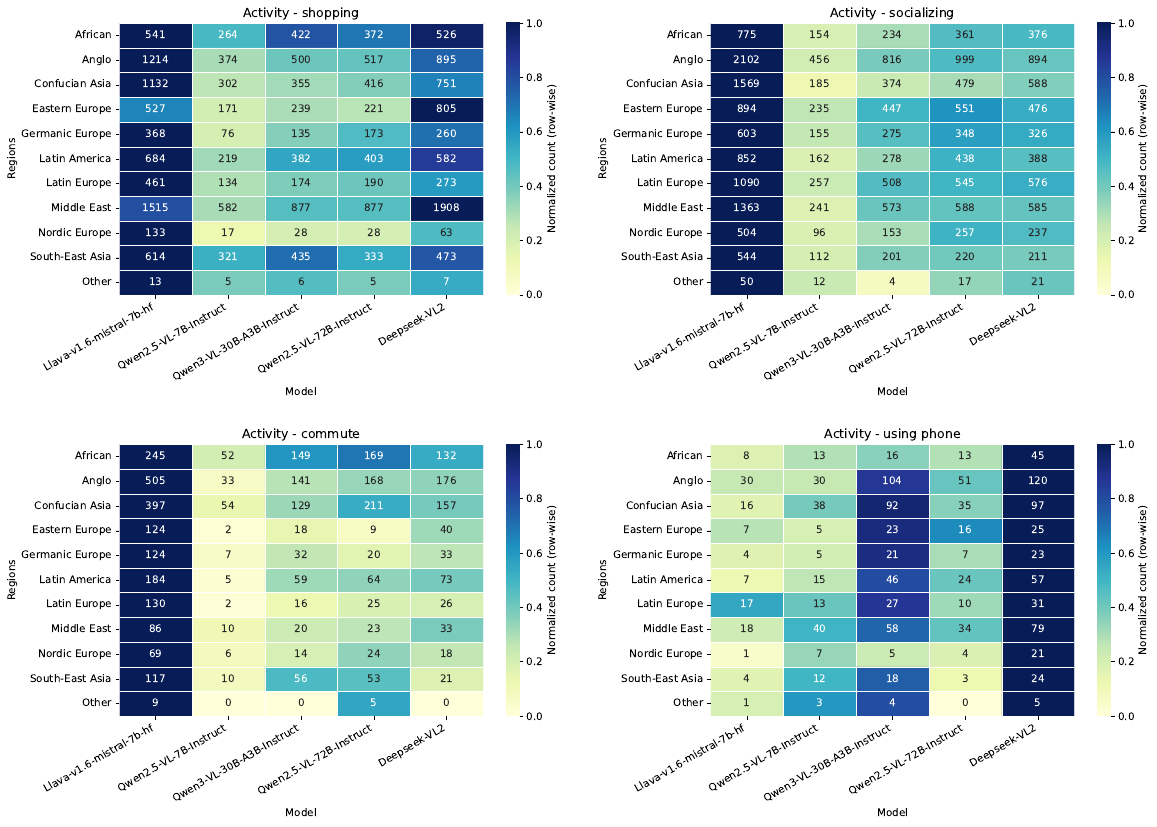}
    \caption{Regional distribution of activity recognition across VLMs for ``commute'', ``using phone'', and ``socializing'' .}
    \label{fig:region_vs_model_4activities_different}
\end{figure}

We continue the discussion on analyzing activities across regions, which was then exemplified with Qwen2.5-VL 72B in Figure \ref{fig:activity_count_by_region}. We now compare across VLMs the counts of activities ``walking'', ``riding motorcycle'', ``moving goods'', and ``cooking''. As shown in Figure \ref{fig:region_vs_model_4activities_similar}, a consensus among the models indicates a low probability of bias. 
For example, the appearance of ``walking'' in all regions and ``cooking'' or ``riding motorcycle'' in South-East Asia, respectively, highlights that activities can be region-specific and model agreement can reveal their presence. Conversely, Figure \ref{fig:region_vs_model_4activities_different} portrays a contrasting scenario with lower agreement. Activities such as ``shopping'' or ``commute'' show significant variance between the models. Notably, LLaVA-v1.6 tends to overestimate the frequencies, which could potentially indicate model bias for certain activities.

\begin{figure}[t!] 
    \centering
    \includegraphics[width=1\linewidth]{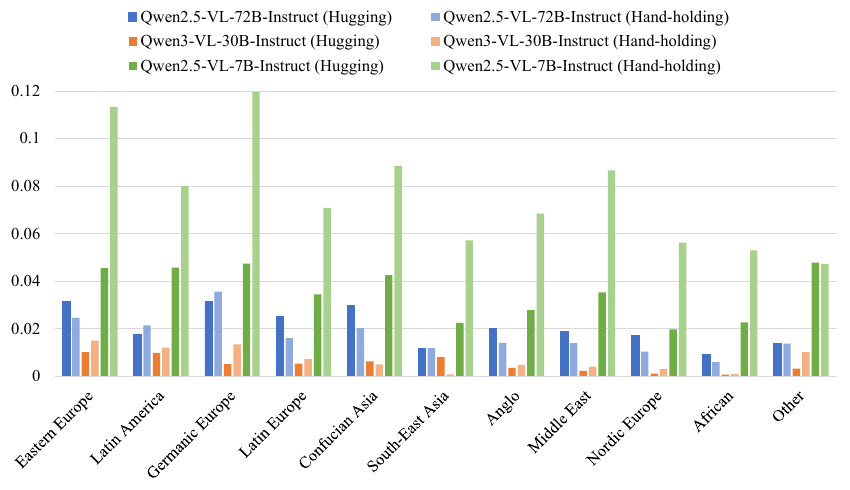}
    \vspace{-0.5cm}
    \caption{Distribution of proximity activities (Hugging and Hand holding) grouped by GLOBE regions. Inside a region, each pair lines represent predictions from a different model.}
    \label{fig:proximity_stats}
\end{figure}

\begin{figure*}[t!]
    \centering
    \begin{subfigure}[b]{0.45\textwidth}
        \centering
        \includegraphics[width=\textwidth]{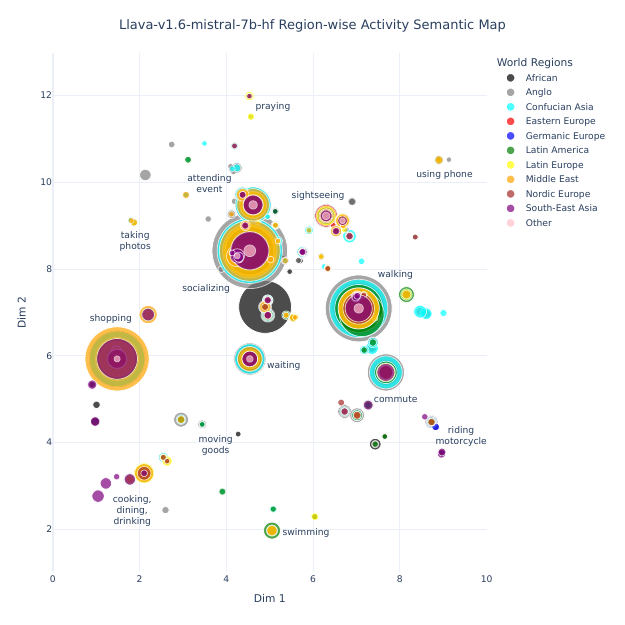}
        \caption{Llava-v1.6}
    \end{subfigure}
    \hfill
    \begin{subfigure}[b]{0.45\textwidth}
        \centering
        \includegraphics[width=\textwidth]{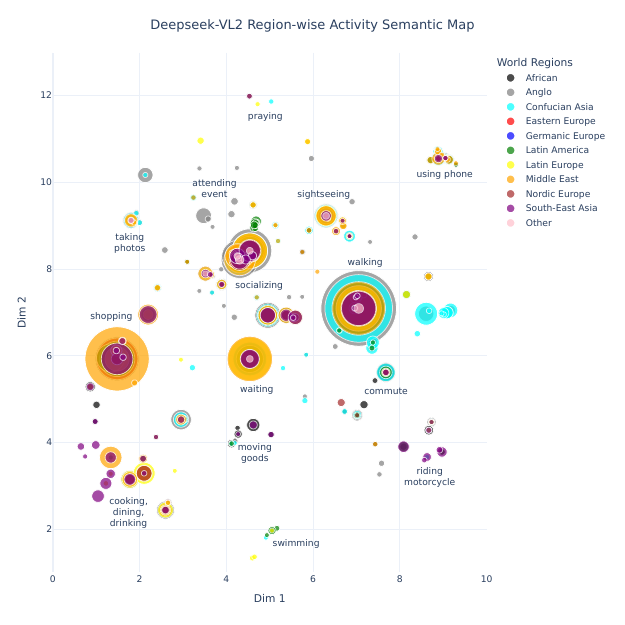}
        \caption{DeepSeek-VL2}
    \end{subfigure}
    \vspace{0.5em}
    \begin{subfigure}[b]{0.45\textwidth}
        \centering
        \includegraphics[width=\textwidth]{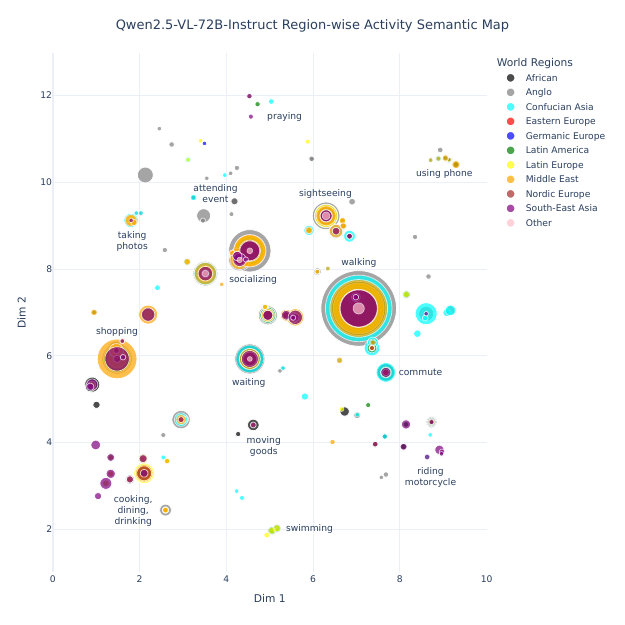}
        \caption{Qwen2.5-72B}
    \end{subfigure}
    \hfill
    \begin{subfigure}[b]{0.45\textwidth}
        \centering
        \includegraphics[width=\textwidth]{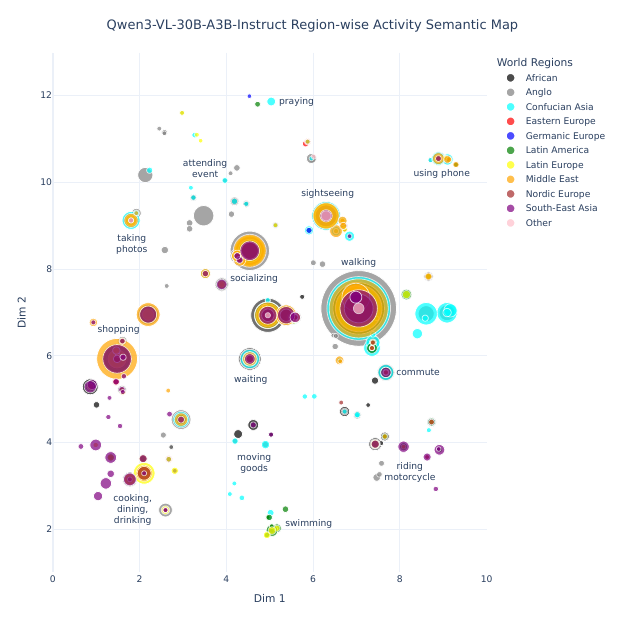}
        \caption{Qwen3-30B}
    \end{subfigure}
    \vspace{-0.2cm}
    \caption{Comparison of Semantic Activity Maps across five different models.}
    \label{fig:model_activity_distribution}
\end{figure*}

Given the open-ended nature of our activity recognition experiments, we also present the results of probing for a specific activity. Figure \ref{fig:proximity_stats} shows the normalized counts of ``Hugging'' and ``Hand-holding'', as a means of understanding proxemics in a region. We obtain these by uniformly sampling clips over countries and regions to ensure fair representation. The counts within samples are normalized by the number of groups and the average of normalized counts across all samples is plotted by region. In the figure, Hand-holding is relatively more common due to its lower-barrier in comparison with Hugging, which generally requires a higher level of intimacy. Regions such as Eastern Europe or Latin America might reflect genuine behavioral trends; however, the African region may not be truly representative. As before, the models differ in their perceived frequencies, indicating different sensitivities; however, the variations across regions are consistent for all models. For example, African has lower frequencies than Latin America.

In addition, we produce Figure \ref{fig:model_activity_distribution} to gain a better perspective of the overall distribution of activities. The plot visualizes embeddings from Qwen3-8B-Embedding by projecting them into a two-dimensional plane with UMAP \cite{mcinnes2020umap}. Each activity in our fixed, data-driven vocabulary is represented by a circle of radius proportional to its frequency and color coded by its region of occurrence. We observe that, in addition to the differences in the distribution of LLaVA-v1.6, the cluster structures of Qwen2.5-VL 72B and DeepSeek-VL2 are near-identical. Such an analysis is beneficial towards identifying parameter-efficient models that can achieve similar performance as the larger models.

\newpage

\section{Additional Experiments}

\textbf{\dataset{}}.
We also include smaller models in the experiments for \dataset{}, such as Cosmos-Reason2 2B, Qwen2.5 (3B–7B), and Qwen3 4B.
 ~\autoref{table:sup_egogroups_crowd_density_ap}
 and ~\autoref{table:sup_egogroups_crowd_density_f1} summarizes the results segmented by crowd density using the AP and F1 score, respectively. Notably, Qwen2.5 (3B–7B) outperforms other models in crowded settings (e.g., F1 scores). 
 ~\autoref{table:sup_egogroups_f1_ap} show overall results in terms of F1 and AP metrics.
 ~\autoref{table:sup_egogroups_globe_regions} reports the AP results segmented by GLOBE region.

\noindent\textbf{JRDB}.
Similarly, ~\autoref{table:sup_results_JRDB_singleframe} includes smaller models for the JRDB dataset. In contrast, these smaller models perform worse than larger model variants.\\

\begin{table*}[h!]
    \centering
    \footnotesize
    \caption{Full fine-grained experiments performance comparison on EgoGroups dataset segmented by
crowd density: scattered, moderate and crowded. Model combination indicates the
backbone family, number of parameters, and modality. G$_{1}$–G$_{5}$ represent AP scores for
group sizes 1–5+, where AP is the average performance across groups.}
    \resizebox{0.99\linewidth}{!}{
    \begin{tabular}{lllcccccccccccccccccc}
    \toprule
        \multirow{2}{*}{Model} &  &  & 
        \multicolumn{6}{c}{\textbf{Scattered}}
        & \multicolumn{6}{c}{\textbf{Moderate}} & \multicolumn{6}{c}{\textbf{Crowded}} \\
        \cmidrule(lr){4-9} \cmidrule(lr){10-15} \cmidrule(lr){16-21}
        & & & G$_{1}$ & G$_{2}$ & G$_{3}$ & G$_{4}$ & G$_{5}$ & AP & G$_{1}$ & G$_{2}$ & G$_{3}$ & G$_{4}$ & G$_{5}$ & AP & G$_{1}$ & G$_{2}$ & G$_{3}$ & G$_{4}$ & G$_{5}$ & AP \\
    \midrule
        \multirow{4}{*}{\rotatebox{10}{\shortstack{Cosmos-\\Reason2}}} & \multirow{2}{*}{2B} & \multirow{1}{*}{VLM} & 76.21 & 59.89 & 61.84 & 54.86 & 45.0 & 59.56 & 58.4 & 51.44 & 55.98 & 48.75 & 57.55 & 54.42 & 54.69 & 46.57 & 49.09 & 49.07 & 48.8 & 49.64 \\
        \addlinespace
         &  & \multirow{1}{*}{LLM}  & 71.01 & 59.22 & 62.99 & 60.42 & 37.12 & 58.15 & 55.68 & 48.81 & 56.9 & 62.92 & 52.45 & 55.35 &  48.73 & 41.51 & 51.45 & 52.77 & 55.81 & 50.05 \\
        \cmidrule(lr){3-21}
         & \multirow{2}{*}{8B} & \multirow{1}{*}{VLM} & 57.18 & 74.42 & 76.36 & 75.35 & 80.6 & 72.78 & 28.92 & 45.28 & 59.54 & 68.02 & 68.55 & 54.06 & 12.8 & 30.12 & 50.05 & 61.53 & 73.18 & 45.53 \\
        \addlinespace
         &  & \multirow{1}{*}{LLM} & 60.72 & 71.03 & 72.84 & 70.49 & 75.0 & 70.02 & 32.47 & 49.61 & 65.81 & 62.54 & 76.02 & 57.29 &  20.58 & 40.56 & 55.9 & 59.69 & 65.15 & 48.38 \\
         \midrule
        \multirow{8}{*}{\rotatebox{10}{\shortstack{Qwen2.5}}} & \multirow{2}{*}{3B} & \multirow{1}{*}{VLM} & 60.06 & 76.93 & 71.16 & 65.63 & 53.64 & 65.48 & 38.75 & 61.15 & 68.92 & 61.92 & 56.49 & 57.45 & 39.46 & 55.47 & 62.98 & 60.16 & 61.66 & 55.94 \\
        \addlinespace
         &  & \multirow{1}{*}{LLM} & 75.64 & 70.24 & 59.69 & 59.03 & 32.88 & 59.5 & 57.69 & 69.58 & 64.59 & 53.65 & 49.75 & 59.06 & 48.45 & 63.46 & 62.94 & 51.52 & 49.29 & 55.13  \\
        \cmidrule(lr){3-21}
         & \multirow{2}{*}{7B} & \multirow{1}{*}{VLM} & \textbf{83.52} & 68.09 & 54.34 & 51.04 & 45.91 & 60.58 &  \textbf{72.85} & 65.86 & 58.96 & 58.38 & 49.01 & 61.01 & \textbf{70.75} & 61.96 & 56.23 & 42.62 & 45.33 & 55.38  \\
        \addlinespace
         &  & \multirow{1}{*}{LLM} & 66.25 & 78.04 & 79.67 & 70.48 & 54.85 & 69.86 & 43.94 & 62.46 & 70.36 & 68.22 & 55.66 & 60.13 & 30.2 & 55.28 & 70.29 & 61.53 & 61.26 & 55.71 \\
        \cmidrule(lr){3-21}
         & \multirow{2}{*}{32B} & \multirow{1}{*}{VLM} & 58.94 & 83.37 & \textbf{86.92} & \textbf{81.6} & \textbf{92.27} & \textbf{80.62} & 30.28 & 62.72 & 77.55 & \textbf{84.21} & 79.38 & 66.83 & 19.11 & 52.42 & 73.26 & 76.05 & 71.56 & 58.48 \\
        \addlinespace
         &  & \multirow{1}{*}{LLM} &  63.71 & 85.28 & 82.38 & 79.51 & 62.73 & 74.72 & 40.49 & 73.14 & 75.28 & 73.02 & 70.62 & 66.51 & 30.45 & 64.66 & 72.69 & 68.27 & 64.38 & 60.09 \\
        \cmidrule(lr){3-21}
         & \multirow{2}{*}{72B} & \multirow{1}{*}{VLM} & 60.53 & 85.1 & 85.11 & 79.17 & 84.54 & 78.89 &  37.18 & 66.91 & \textbf{79.14} & 78.45 & \textbf{82.76} & \textbf{68.89} & 25.89 & 59.79 & \textbf{73.79} & \textbf{77.81} & \textbf{74.18} & \textbf{62.29} \\
        \addlinespace
         &  & \multirow{1}{*}{LLM} & 63.97 & \textbf{85.84} & 80.24 & 75.0 & 59.55 & 72.92 & 38.73 & 73.68 & 76.4 & 71.85 & 69.28 & 65.99 & 28.02 & 65.93 & 73.18 & 68.64 & 61.92 & 59.54  \\
         \midrule
        \multirow{4}{*}{\rotatebox{10}{\shortstack{Qwen3}}} & \multirow{2}{*}{4B} & \multirow{1}{*}{VLM} & 82.16 & 68.09 & 70.03 & 71.88 & 76.97 & 73.83 & 50.47 & 55.72 & 61.93 & 70.38 & 64.93 & 60.69 & 32.31 & 44.12 & 58.22 & 67.86 & 68.47 & 54.2\\
        \addlinespace
         &  & \multirow{1}{*}{LLM}  & 68.28 & 78.47 & 80.45 & 72.57 & 68.94 & 73.74 & 46.76 & 63.03 & 72.77 & 77.22 & 73.2 & 66.6 &  35.84 & 52.57 & 70.46 & 70.62 & 67.86 & 59.47\\
        \cmidrule(lr){3-21}
         & \multirow{2}{*}{30B} & \multirow{1}{*}{VLM} &  71.59 & 80.28 & 74.26 & 75.0 & 59.09 & 72.04 & 50.29 & 66.02 & 71.33 & 71.07 & 70.36 & 65.81 & 36.89 & 57.26 & 63.7 & 60.82 & 58.39 & 55.41 \\
        \addlinespace
         &  & \multirow{1}{*}{LLM} & 69.41 & 83.56 & 79.37 & 78.47 & 62.73 & 74.71 & 54.94 & 69.94 & 71.17 & 72.14 & 70.15 & 67.67 &  47.59 & 62.84 & 69.35 & 63.25 & 59.25 & 60.46 \\
         \midrule
        \multirow{2}{*}{\rotatebox{10}{\shortstack{Gemini-\\3-Pro}}} &  & \multirow{1}{*}{VLM}  & 82.67 & 80.85 & 73.92 & 76.39 & 68.79 & 76.52 & 70.37 & \textbf{75.5} & 68.24 & 67.24 & 57.07 & 67.69 & 63.09 & \textbf{69.78} & 64.16 & 54.43 & 47.9 & 59.87  \\
        \addlinespace
         &  & \multirow{1}{*}{LLM} & 77.66 & 81.85 & 75.33 & 75.35 & 59.55 & 73.95 & 60.29 & 74.18 & 73.23 & 70.9 & 62.12 & 68.14 & 50.95 & 69.0 & 68.63 & 63.71 & 54.88 & 61.44 \\
    \bottomrule
    \end{tabular}
    }
    \label{table:sup_egogroups_crowd_density_ap}
\end{table*}

\begin{table*}[h!]
    \centering
    \footnotesize
    \caption{Full fine-grained experiments performance comparison on EgoGroups dataset segmented by
crowd density: scattered, moderate and crowded. Model combination indicates the
backbone family, number of parameters, and modality. Precision, Recall, and F1 scores
measure the false positive rate of predicted group IDs.}
    \resizebox{0.65\linewidth}{!}{
    \begin{tabular}{lllccccccccc}
    \toprule
        \multirow{2}{*}{Model} &  & 
        & \multicolumn{3}{c}{\textbf{Scattered}}
        & \multicolumn{3}{c}{\textbf{Moderate}} & \multicolumn{3}{c}{\textbf{Crowded}} \\
        \cmidrule(lr){4-6} \cmidrule(lr){7-9} \cmidrule(lr){10-12}
        & & & Precision & Recall & F1 & Precision & Recall & F1   & Precision & Recall & F1  \\
    \midrule
        \multirow{4}{*}{\rotatebox{10}{\shortstack{Cosmos-\\Reason2}}} & \multirow{2}{*}{2B} & \multirow{1}{*}{VLM} & 49.03 & 25.24 & 33.29 & 33.97 & 17.3 & 22.88 & 25.88 & 12.77 & 17.05 \\
        \addlinespace
         &  & \multirow{1}{*}{LLM} & 38.24 & 21.5 & 27.52 & 32.46 & 17.74 & 22.92 & 26.56 & 14.07 & 18.37  \\
        \cmidrule(lr){3-12}
         & \multirow{2}{*}{8B} & \multirow{1}{*}{VLM}  & 50.44 & 49.82 & 50.1 & 25.92 & 16.16 & 19.9 &  11.22 & 5.41 & 7.3 \\
        \addlinespace
         &  & \multirow{1}{*}{LLM}  & 44.73 & 40.6 & 42.47 &  22.04 & 14.3 & 17.33 & 11.75 & 6.61 & 8.45  \\
        \midrule
        \multirow{8}{*}{\rotatebox{10}{\shortstack{Qwen2.5}}} & \multirow{2}{*}{3B} & \multirow{1}{*}{VLM}  & 46.41 & 54.8 & 50.23 &  32.77 & 33.8 & 33.25 &  26.81 & 29.27 & \textbf{27.98} \\
        \addlinespace
         &  & \multirow{1}{*}{LLM}  & 41.69 & 40.13 & 40.81 & 34.29 & \textbf{39.04} & 36.45 & 23.09 & \textbf{30.58} & 26.3  \\
        \cmidrule(lr){3-12}
         & \multirow{2}{*}{7B} & \multirow{1}{*}{VLM}  & 47.95 & 34.0 & 39.66 & \textbf{39.65} & 32.59 & 35.66 & \textbf{28.2} & 26.98 & 27.57 \\
        \addlinespace
         &  & \multirow{1}{*}{LLM} & 47.76 & 53.36 & 50.4 & 26.32 & 22.7 & 24.37 &  17.23 & 15.5 & 16.31 \\
        \cmidrule(lr){3-12}
         & \multirow{2}{*}{32B} & \multirow{1}{*}{VLM}  & 51.4 & 61.87 & 56.12 &  27.05 & 26.18 & 26.58 &  14.35 & 13.69 & 14.01  \\
        \addlinespace
         &  & \multirow{1}{*}{LLM} & 51.22 & 64.05 & 56.88 & 29.8 & 33.25 & 31.41 & 17.5 & 21.74 & 19.38\\
        \cmidrule(lr){3-12}
         & \multirow{2}{*}{72B} & \multirow{1}{*}{VLM}  & 52.77 & \textbf{65.65} & 58.49 & 32.76 & 33.54 & 33.1 &  21.0 & 22.25 & 21.6 \\
        \addlinespace
         &  & \multirow{1}{*}{LLM}  & 51.09 & 65.36 & 57.3 & 28.52 & 32.39 & 30.31 &  15.01 & 18.98 & 16.75  \\
        \midrule
        \multirow{4}{*}{\rotatebox{10}{\shortstack{Qwen3}}} & \multirow{2}{*}{4B} & \multirow{1}{*}{VLM}  & 60.6 & 39.35 & 47.53 & 27.85 & 17.7 & 21.63 & 13.93 & 8.43 & 10.5 \\
        \addlinespace
         &  & \multirow{1}{*}{LLM}  & 53.27 & 52.36 & 52.71 & 29.78 & 25.1 & 27.21 & 13.65 & 11.77 & 12.63  \\
        \cmidrule(lr){3-12}
         & \multirow{2}{*}{30B} & \multirow{1}{*}{VLM}  & 54.88 & 56.73 & 55.7 & 30.02 & 25.69 & 27.68 & 13.73 & 12.45 & 13.06  \\
        \addlinespace
         &  & \multirow{1}{*}{LLM}  & 55.96 & 61.84 & 58.73 & 33.72 & 29.54 & 31.47 & 17.92 & 16.29 & 17.03 \\
         \midrule
        \multirow{2}{*}{\rotatebox{10}{\shortstack{Gemini-\\3-Pro}}} & & \multirow{1}{*}{VLM} & \textbf{61.02} & 57.83 & \textbf{59.19} & 38.44 & 36.99 & \textbf{37.63} & 22.75 & 25.67 & 24.09 \\
        \addlinespace
         &  & \multirow{1}{*}{LLM}  & 58.0 & 59.01 & 58.26 & 34.59 & 34.21 & 34.34 & 19.28 & 22.72 & 20.83
         \\
    \bottomrule
    \end{tabular}
    }
    \label{table:sup_egogroups_crowd_density_f1}
\end{table*}

\begin{table*}[h!]
    \centering
    \footnotesize
    \caption{Full fine-grained experiments performance comparison on the \dataset{} dataset. Model combination indicates the backbone family, number of parameters, and input modality. G$_{1}$–G$_{5}$
represent AP scores for group sizes 1–5+, with AP showing the average performance
across groups. Precision, Recall, and F1 scores measure the false positive rate of
predicted group IDs.}
    \resizebox{0.65\linewidth}{!}{
    \begin{tabular}{lllccccccccc}
    \toprule
        \multirow{1}{*}{Model} &  & & G$_{1}$ & G$_{2}$ & G$_{3}$ & G$_{4}$ & G$_{5}$ & AP & Precision & Recall & F1 \\
    \midrule
        \multirow{4}{*}{\rotatebox{10}{\shortstack{Cosmos-\\Reason2}}} & \multirow{2}{*}{2B} & \multirow{1}{*}{VLM} & 57.49 & 49.88 & 52.72 & 49.29 & 51.65 & 52.21 & 31.37 & 15.75 & 20.96 \\
        \addlinespace
         &  & \multirow{1}{*}{LLM} & 52.55 & 46.24 & 54.35 & 57.03 & 53.26 & 52.68 & 30.07 & 14.01 & 17.64 \\
        \cmidrule(lr){3-12}
         & \multirow{2}{*}{8B} & \multirow{1}{*}{VLM}  & 21.09 & 40.74 & 55.68 & 65.4 & 72.46 & 51.07 & 23.83 & 14.01 & 17.64 \\
        \addlinespace
         &  & \multirow{1}{*}{LLM} & 27.27 & 47.32 & 60.96 & 61.9 & 69.47 & 53.38 & 20.59 & 12.99 & 15.93 \\
         \midrule
        \multirow{8}{*}{\rotatebox{10}{\shortstack{Qwen2.5}}} & \multirow{2}{*}{3B} & \multirow{1}{*}{VLM} & 40.9 & 59.96 & 65.69 & 61.21 & 59.11 & 57.37 & 31.16 & 33.56 & 32.32 \\
        \addlinespace
         &  & \multirow{1}{*}{LLM} & 53.35 & 66.57 & 63.23 & 52.78 & 48.08 & 56.8 & 28.58 & \textbf{34.78} & 31.35 \\
        \cmidrule(lr){3-12}
         & \multirow{2}{*}{7B} & \multirow{1}{*}{VLM} & \textbf{72.39} & 64.12 & 57.06 & 49.02 & 46.27 & 57.77 & \textbf{33.83} & 29.82 & 31.69  \\
        \addlinespace
         &  & \multirow{1}{*}{LLM} & 37.12 & 60.51 & 71.1 & 64.68 & 58.91 & 58.47 & 24.34 & 22.08 & 23.14 \\
        \cmidrule(lr){3-12}
         & \multirow{2}{*}{32B} & \multirow{1}{*}{VLM}  & 25.59 & 59.72 & 75.94 & \textbf{79.67} & 75.91 & 63.37 & 23.7 & 23.32 & 23.49 \\
        \addlinespace
         &  & \multirow{1}{*}{LLM} & 36.06 & 70.15 & 74.43 & 71.16 & 66.38 & 63.64 & 25.45 & 30.41 & 27.7 \\
        \cmidrule(lr){3-12}
         & \multirow{2}{*}{72B} & \multirow{1}{*}{VLM}  & 31.96 & 65.24 & \textbf{76.65} & 78.27 & \textbf{77.86} & \textbf{66.0} & 20.09 & 30.95 & 29.97 \\
        \addlinespace
         &  & \multirow{1}{*}{LLM} & 34.04 & 71.03 & 74.9 & 70.46 & 64.22 & 62.93 & 23.68 & 28.78 & 25.97 \\
         \midrule
        \multirow{4}{*}{\rotatebox{10}{\shortstack{Qwen3}}} & \multirow{2}{*}{4B} & \multirow{1}{*}{VLM}  & 41.65 & 51.16 & 60.68 & 69.21 & 68.06 & 58.15 & 24.29 & 15.09 & 18.61  \\
        \addlinespace
         &  & \multirow{1}{*}{LLM} & 41.64 & 59.39 & 72.18 & 73.22 & 69.79 & 63.24 & 24.18 & 20.97 & 22.44 \\
        \cmidrule(lr){3-12}
         & \multirow{2}{*}{30B} & \multirow{1}{*}{VLM}  & 43.61 & 63.1 & 67.4 & 65.82 & 62.65 & 60.51 & 24.40 & 21.96 & 23.11  \\
        \addlinespace
         &  & \multirow{1}{*}{LLM} & 51.49 & 67.81 & 70.87 & 67.8 & 63.2 & 64.23 & 28.34 & 25.94 & 27.05 \\
         \midrule
        \multirow{2}{*}{\rotatebox{10}{\shortstack{Gemini-\\3-Pro}}} &  & \multirow{1}{*}{VLM} & 66.8 & \textbf{73.18} & 66.34 & 61.16 & 52.79 & 64.05 & 31.76 & 33.25 & \textbf{32.44} \\
        \addlinespace
         &  & \multirow{1}{*}{LLM}  & 55.82 & 72.38 & 70.79 & 67.47 & 57.81 & 64.85 & 28.24 & 30.79 & 29.41 \\
    \bottomrule
    \end{tabular}
    }
    \label{table:sup_egogroups_f1_ap}

\end{table*}

\begin{table*}[h!]
    \centering
    \footnotesize
    \caption{Full fine-grained experiments performance comparison on the EgoGroups dataset, segmented
by GLOBE regions. Model combination
indicates the backbone family, number of parameters, and input modality. We report
each region's AP score individually, and compute MAP as the average across all regions.
The best and worst values per row are highlighted in red and blue, respectively.}
    \resizebox{0.6\linewidth}{!}{
    \begin{tabular}{llllcccccccccc}
    \toprule
        \multirow{1}{*}{Model} & & 
        & \textbf{AF}
        & \textbf{AN} & \textbf{CA} & \textbf{EU} & \textbf{GE} & \textbf{LA} & \textbf{LE} & \textbf{ME} &\textbf{NE} & \textbf{SA} & \textbf{O} \\
    \midrule
        \multirow{4}{*}{\rotatebox{10}{\shortstack{Cosmos-\\Reason2}}} & \multirow{2}{*}{2B} & \multirow{1}{*}{VLM}  & \textcolor{red}{57.13} & 53.18 & 53.67 & 56.16 & \textcolor{blue}{44.94} & 47.93 & 51.12 & 49.50 & 46.30 & 50.67 & 50.64  \\
        \addlinespace
         &  & \multirow{1}{*}{LLM}   & \textcolor{blue}{48.85} & 51.74 & 54.61 & 53.88 & 51.64 & \textcolor{red}{58.28} & 49.75 & 50.54 & 57.31 & 53.96 & 53.51 \\
        \cmidrule(lr){3-14}
         & \multirow{2}{*}{8B} & \multirow{1}{*}{VLM} & 47.19 & 50.49 & 54.01 & 51.69 & \textcolor{blue}{40.07} & 51.12 & 52.98 & 47.05 & 49.86 & 52.86 & \textcolor{red}{58.89}  \\
        \addlinespace
         &  & \multirow{1}{*}{LLM} & \textcolor{blue}{45.95} & \textcolor{red}{56.89} & 55.25 & 54.59 & 50.95 & 54.17 & 54.40 & 49.18 & 51.21 & 52.95 & 53.37 \\
         \midrule
        \multirow{8}{*}{\rotatebox{10}{\shortstack{Qwen2.5}}} & \multirow{2}{*}{3B} & \multirow{1}{*}{VLM} & 55.45 & 58.24 & 58.29 & 58.09 & 56.34 & \textcolor{blue}{54.06} & 55.75 & 55.14 & \textcolor{red}{60.74} & 56.32 & 54.53 \\
        \addlinespace
         &  & \multirow{1}{*}{LLM} & 55.86 & 58.69 & 55.46 & 56.71 & \textcolor{red}{59.75} & 57.48 & 54.51 & \textcolor{blue}{53.88} & 57.93 & 56.55 & 57.71 \\
        \cmidrule(lr){3-14}
         & \multirow{2}{*}{7B} & \multirow{1}{*}{VLM} & 54.33 & 58.52 & 57.26 & 62.57 & 61.81 & \textcolor{red}{65.47} & 57.05 & \textcolor{blue}{50.81} & 59.05 & 63.33 & 56.34 \\
        \addlinespace
         &  & \multirow{1}{*}{LLM} & 55.70 & 60.27 & 57.77 & \textcolor{blue}{52.77} & 61.74 & \textcolor{red}{63.10} & 59.89 & 54.37 & 61.89 & 60.01 & 58.42  \\
        \cmidrule(lr){3-14}
         & \multirow{2}{*}{32B} & \multirow{1}{*}{VLM} & 61.99 & 64.63 & 63.57 & 58.82 & 60.48 & 66.55 & 64.09 & 57.98 & \textcolor{red}{71.75} & 65.75 & \textcolor{blue}{50.00} \\
        \addlinespace
         &  & \multirow{1}{*}{LLM} & 61.87 & 65.01 & 62.16 & 61.69 & 62.84 & 66.89 & 63.84 & \textcolor{blue}{59.75} & \textcolor{red}{69.36} & 63.28 & 61.63 \\
        \cmidrule(lr){3-14}
         & \multirow{2}{*}{72B} & \multirow{1}{*}{VLM} & \textcolor{blue}{58.64} & 67.24 & 65.78 & 64.54 & 67.06 & 70.06 & 67.69 & 59.42 & \textcolor{red}{71.45} & 67.30 & 65.80 \\
        \addlinespace
         &  & \multirow{1}{*}{LLM} & 61.94 & 65.22 & 61.88 & 62.32 & 64.28 & 65.04 & 62.61 & 57.33 & \textcolor{red}{67.31} & 64.49 & \textcolor{blue}{54.21} \\
         \midrule
        \multirow{4}{*}{\rotatebox{10}{\shortstack{Qwen3}}} & \multirow{2}{*}{4B} & \multirow{1}{*}{VLM} & 60.17 & 58.78 & 60.25 & 59.91 & 61.57 & \textcolor{blue}{53.04} & 56.21 & 56.47 & 53.61 & 53.53 & \textcolor{red}{61.83}\\
        \addlinespace
         &  & \multirow{1}{*}{LLM} & 58.31 & 65.44 & 62.48 & 64.49 & 60.82 & \textcolor{red}{69.47} & 65.44 & 57.79 & 65.56 & 62.83 & \textcolor{blue}{52.35}  \\
        \cmidrule(lr){3-14}
         & \multirow{2}{*}{30B} & \multirow{1}{*}{VLM} & 56.07 & 64.76 & 58.49 & 60.26 & \textcolor{red}{65.85} & 59.92 & 63.42 & \textcolor{blue}{53.94} & 56.30 & 64.20 & 63.10 \\
        \addlinespace
         &  & \multirow{1}{*}{LLM} & 62.64 & 65.61 & 64.43 & 65.95 & 66.34 & 66.73 & 63.30 & 57.70 & \textcolor{red}{67.66} & 66.31 & \textcolor{blue}{55.60}  \\
         \midrule
        \multirow{2}{*}{\rotatebox{10}{\shortstack{Gemini-\\3-Pro}}} & \multirow{2}{*}{} & \multirow{1}{*}{VLM} & 62.41 & 66.70 & 64.01 & 68.23 & 65.53 & 64.79 & 64.32 & \textcolor{blue}{60.24} & \textcolor{red}{71.84} & 66.03 & 62.31  \\
        \addlinespace
         &  & \multirow{1}{*}{LLM} & 60.87 & 64.59 & 64.69 & \textcolor{red}{68.61} & 64.37 & 62.55 & 66.47 & \textcolor{blue}{59.51} & 62.44 & 66.69 & 64.70 \\
    \bottomrule
    \end{tabular}
    }
    \label{table:sup_egogroups_globe_regions}
\end{table*}

\begin{table*}[h!]
    \centering
    \footnotesize
    \caption{Fine-grained performance comparison on the JRDB dataset. Model combination indicates the backbone family, number of parameters, and input modality. $G_1$--$G_5$ represent AP scores for group sizes 1--5$^+$, with AP showing the average performance across all groups. Precision, Recall, and F1 scores measure the false positive rate of predicted group IDs.} 
    \resizebox{0.65\linewidth}{!}{
    \begin{tabular}{lllccccccccc}
    \toprule
        \multirow{2}{*}{Model} & & &
        \multicolumn{6}{c}{\textbf{Group Detection (AP)}} &
        \multicolumn{3}{c}{\textbf{Group ID Prediction}} \\
        \cmidrule(lr){4-9} \cmidrule(lr){10-12}
        & & & G$_{1}$ & G$_{2}$ & G$_{3}$ & G$_{4}$ & G$_{5}$ & AP  & Precision & Recall & F1 \\
    \midrule
    \rowcolor{gray!20}
         \multicolumn{2}{l}{JLSG~\cite{eccv_groups}} & & 8.0 & 29.30 & 37.5 & 65.40 & 67.0 & 41.4 & - & - & - \\
    \rowcolor{gray!20}
         \multicolumn{2}{l}{JRDB-Act~\cite{JRDB-Act}} & & 81.40 & 64.80 & 49.10 & 63.20 & 37.20 & 59.2 & - & - & - \\
    \rowcolor{gray!20}              \multicolumn{2}{l}{DVT3~\cite{Yokoyama_2025_ICCV}} &    & - & - & - & - & - & - &  \textbf{61.16} & 31.06 & 41.19 \\
    \midrule
        \multirow{4}{*}{\rotatebox{10}{\shortstack{Cosmos-\\Reason2}}}
        
         & \multirow{2}{*}{2B} & \multirow{1}{*}{VLM} &
          75.59 & 45.6 & 33.17 & 32.3 & 28.3 & 42.99 & 16.41 & 2.33 & 4.09 \\
         &  & \multirow{1}{*}{LLM} & 51.93 & 39.46 & 36.07 & 36.91 & 42.05 & 41.28 & 17.06 & 3.49 & 5.79 \\
         \cmidrule(lr){2-12}
         & \multirow{2}{*}{8B} & \multirow{1}{*}{VLM} &
          44.88 & 34.7 & 30.96 & 36.83 & 55.0 & 40.47 & 19.83 & 5.83 & 9.02 \\
         &  & \multirow{1}{*}{LLM} & 48.47 & 36.19 & 29.05 & 31.71 & 56.26 & 40.34 & 24.7 & 6.93 & 10.82 \\
        \midrule
        \multirow{4}{*}{\rotatebox{10}{\shortstack{Qwen2.5}}}
        & \multirow{2}{*}{3B} & \multirow{1}{*}{VLM} &
          65.99 & 51.13 & 42.73 & 37.84 & 30.3 & 45.6 & 12.46 & 4.82 & 6.96 \\
         &  & \multirow{1}{*}{LLM} & 55.79 & 45.63 & 36.24 & 35.15 & 47.36 & 44.03 & 11.87 & 4.07 & 6.07 \\
        \cmidrule(lr){2-12}
        & \multirow{2}{*}{7B} & \multirow{1}{*}{VLM} &
          \textbf{79.02} & 50.55 & 38.56 & 32.05 & 24.07 & 44.89 & 12.54 & 7.28 & 9.21 \\
         &  & \multirow{1}{*}{LLM} & 33.58 & 40.37 & 36.47 & 39.93 & 69.4 & 43.95 & 12.56 & 4.92 & 7.07 \\
        \cmidrule(lr){2-12}
        
         & \multirow{2}{*}{32B} & \multirow{1}{*}{VLM} &
          25.54 & 47.44 & 47.23 & 54.03 & 69.3 & 48.71 & 22.39 & 16.81 & 19.2 \\
         &  & \multirow{1}{*}{LLM} & 35.13 & 56.78 & 54.11 & 56.96 & 69.49 & 54.49 & 28.45 & 23.88 & 25.97 \\
        \cmidrule(lr){2-12}
         & \multirow{2}{*}{72B} & \multirow{1}{*}{VLM} &
          28.4 & 45.44 & 43.62 & 51.26 & 71.99 & 48.14 & 23.54 & 17.28 & 19.93 \\
         & & \multirow{1}{*}{LLM} & 32.95 & 55.42 & 55.34 & 60.99 & 66.86 & 54.31 & 28.98 & 25.46 & 27.11 \\
        \midrule
        \multirow{6}{*}{\rotatebox{10}{Qwen3}}
        
         & \multirow{2}{*}{4B} & \multirow{1}{*}{VLM} &
         48.95 & 38.65 & 32.13 & 34.48 & 57.97 & 42.44 & 16.64 & 4.56 & 7.16 \\
         &  & \multirow{1}{*}{LLM} & 39.54 & 44.46 & 43.99 & 56.21 & 64.12 & 49.66 & 23.58 & 11.09 & 15.08 \\
         \cmidrule(lr){2-12}
         & \multirow{2}{*}{30B} & \multirow{1}{*}{VLM} &
         30.5 & 44.2 & 41.5 & 44.71 & 72.48 & 46.68 & 30.19 & 17.45 & 22.12 \\
         &  & \multirow{1}{*}{LLM} & 47.15 & 57.38 & 54.45 & 57.97 & 66.62 & 56.71 & 40.25 & 27.84 & 32.91 \\
        \cmidrule(lr){2-12}
         & \multirow{2}{*}{235B} & \multirow{1}{*}{VLM} & 31.54 & 44.28 & 41.36 & 50.59 & 77.24 & 49.0 & 36.80 & 21.51 & 27.15 \\
         &  & \multirow{1}{*}{LLM} & 31.0 & 51.29 & 60.4 & 69.63 & \textbf{79.95} & 58.46 & 34.03 & 25.53 & 29.17 \\
        \midrule
        \multirow{2}{*}{\rotatebox{10}{\shortstack{Gemini-\\3-Pro}}} &
        & \multirow{1}{*}{VLM} & 56.66 & \textbf{71.79} & 79.33 & 79.7 & 57.49 & 68.99 & 42.99 & \textbf{46.38} & \textbf{44.62} \\
        &  & \multirow{1}{*}{LLM} & 51.65 & 69.97 & \textbf{82.52} & \textbf{81.8} & 64.08 & \textbf{70.0} & 40.89 & 43.97 & 42.37 \\
    \bottomrule
    \end{tabular}
    }
    \label{table:sup_results_JRDB_singleframe}
\end{table*}

\clearpage
\newpage

\section{More Qualitative Examples}

\begin{figure*}[!ht]
\centering

\begin{subfigure}{0.32\linewidth}
\centering
\includegraphics[width=\linewidth]{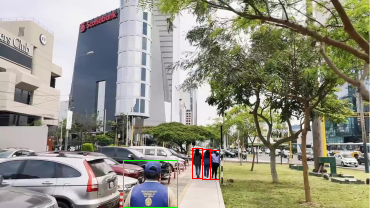}
\end{subfigure}
\hfill
\begin{subfigure}{0.32\linewidth}
\centering
\includegraphics[width=\linewidth]{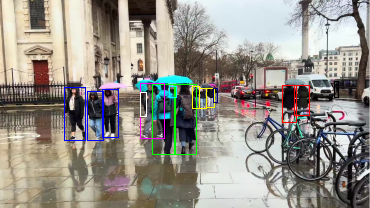}
\end{subfigure}
\hfill
\begin{subfigure}{0.32\linewidth}
\centering
\includegraphics[width=\linewidth]{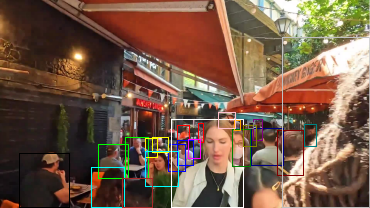}
\end{subfigure}

\medskip

\begin{subfigure}{0.32\linewidth}
\centering
\includegraphics[width=\linewidth]{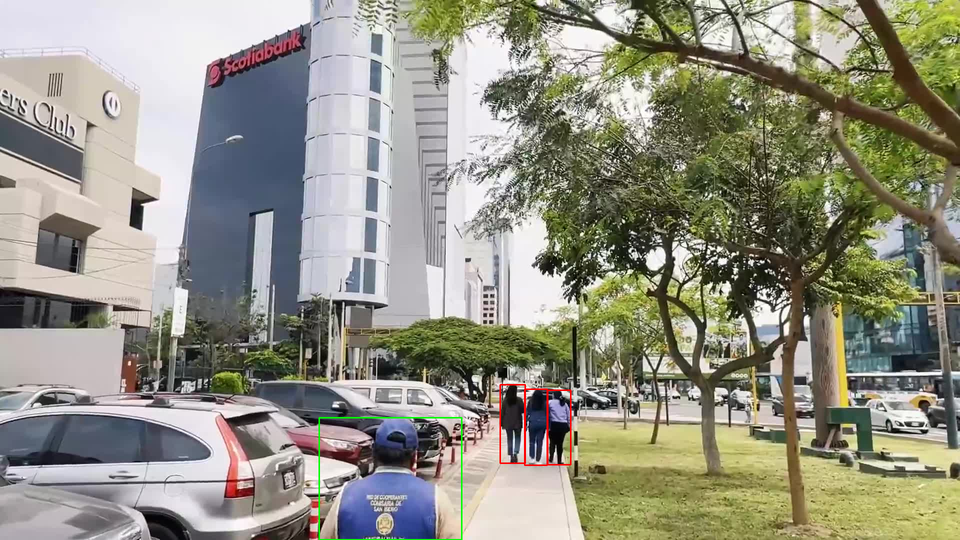}
\end{subfigure}
\hfill
\begin{subfigure}{0.32\linewidth}
\centering
\includegraphics[width=\linewidth]{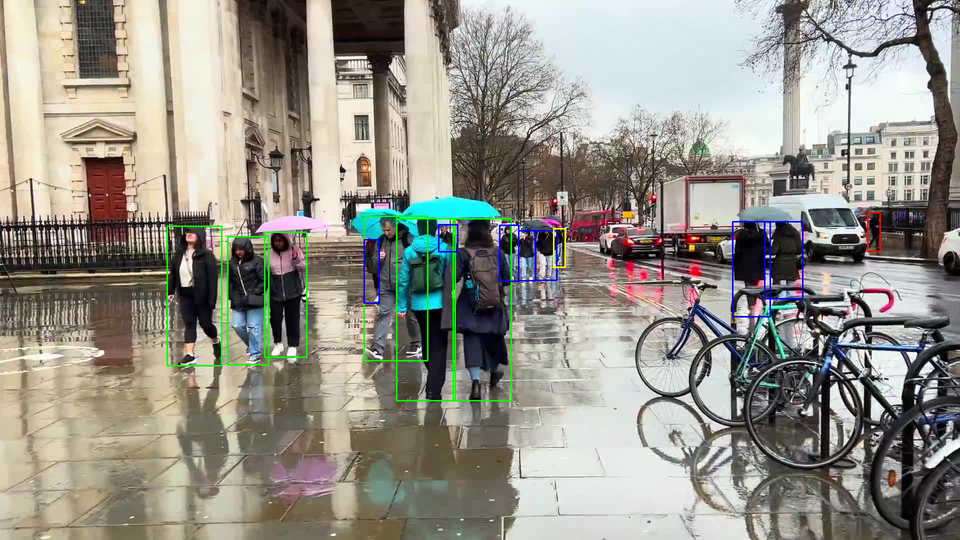}
\end{subfigure}
\hfill
\begin{subfigure}{0.32\linewidth}
\centering
\includegraphics[width=\linewidth]{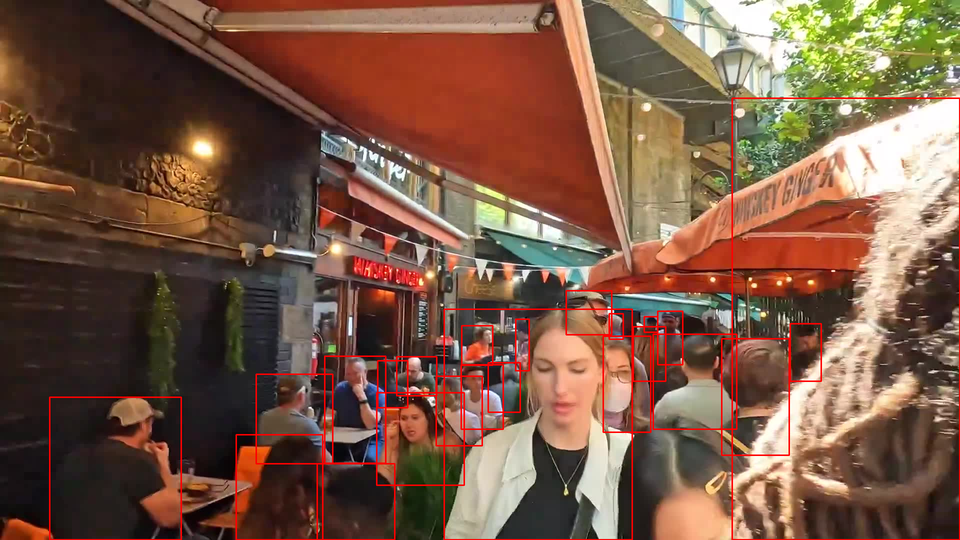}
\end{subfigure}

\medskip

\begin{subfigure}{0.32\linewidth}
\centering
\includegraphics[width=\linewidth]{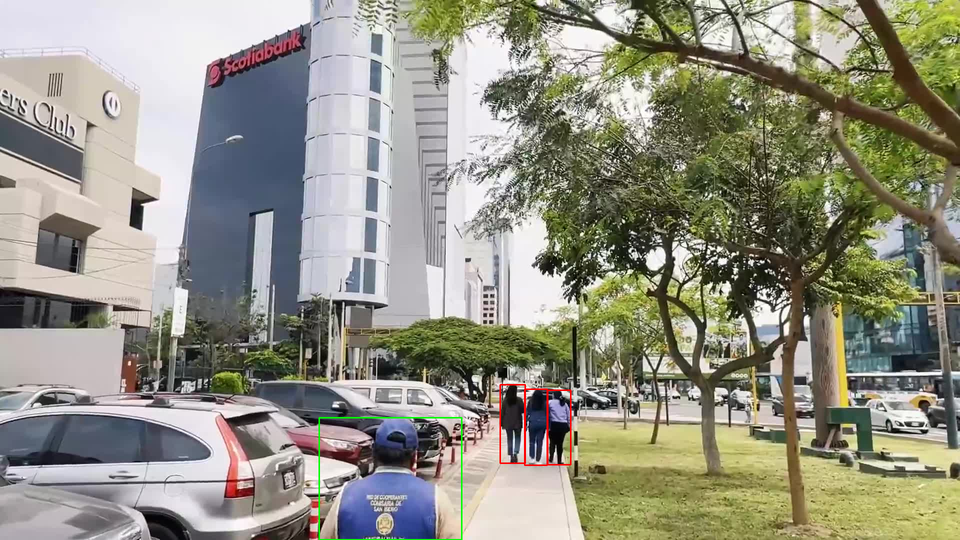}
\end{subfigure}
\hfill
\begin{subfigure}{0.32\linewidth}
\centering
\includegraphics[width=\linewidth]{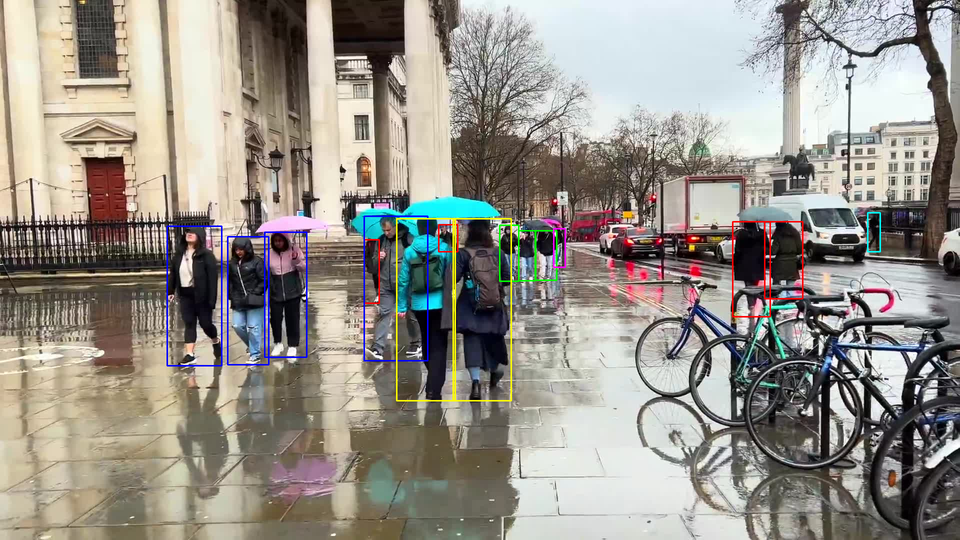}
\end{subfigure}
\hfill
\begin{subfigure}{0.32\linewidth}
\centering
\includegraphics[width=\linewidth]{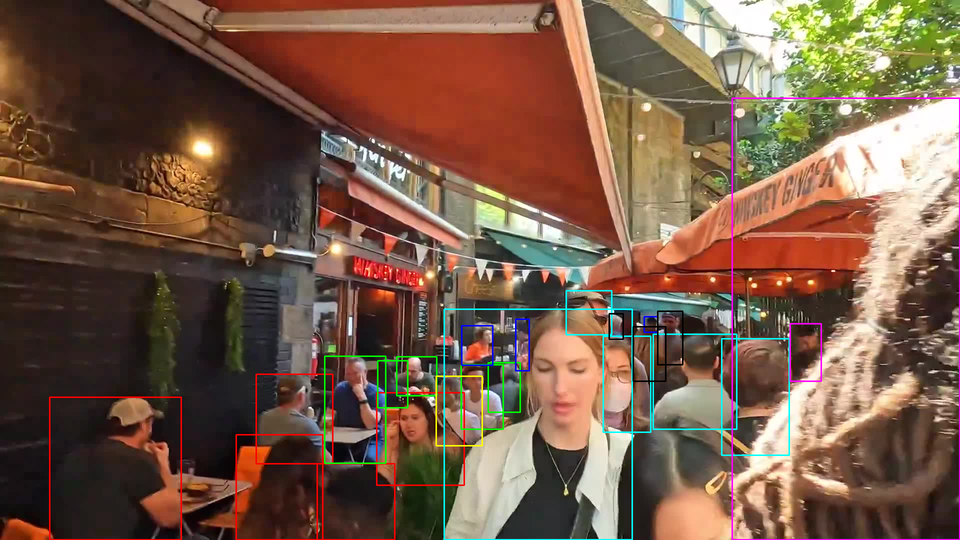}
\end{subfigure}

\medskip

\begin{subfigure}{0.32\linewidth}
\centering
\includegraphics[width=\linewidth]{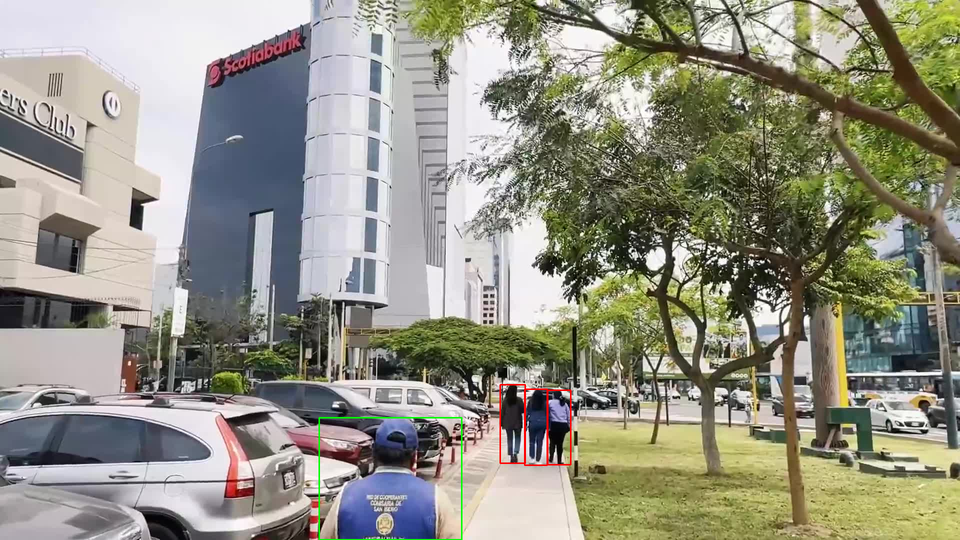}
\end{subfigure}
\hfill
\begin{subfigure}{0.32\linewidth}
\centering
\includegraphics[width=\linewidth]{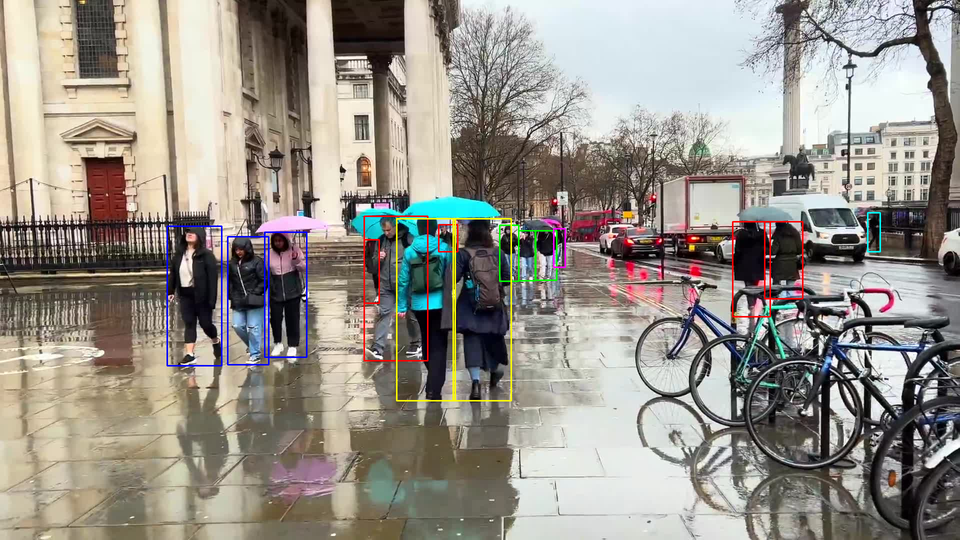}
\end{subfigure}
\hfill
\begin{subfigure}{0.32\linewidth}
\centering
\includegraphics[width=\linewidth]{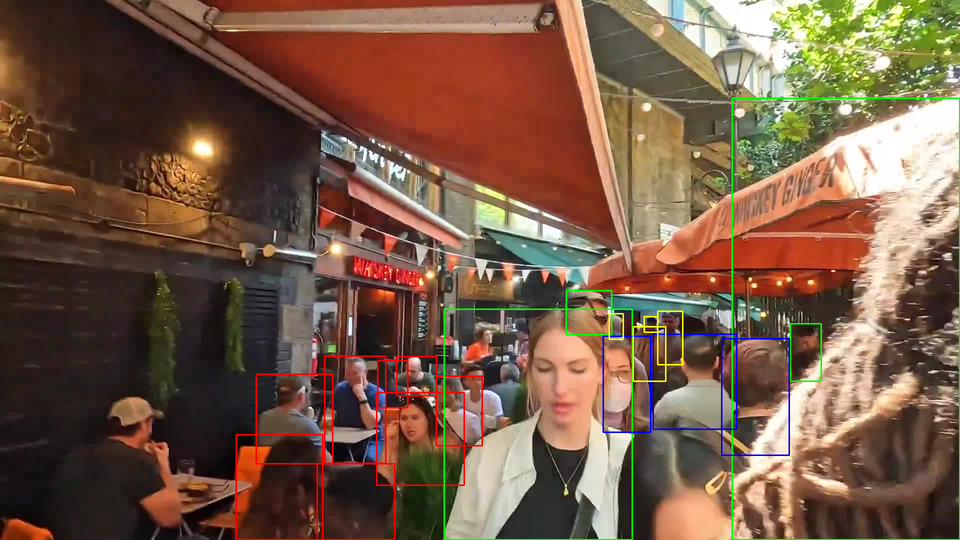}
\end{subfigure}

\medskip

\begin{subfigure}{0.32\linewidth}
\centering
\includegraphics[width=\linewidth]{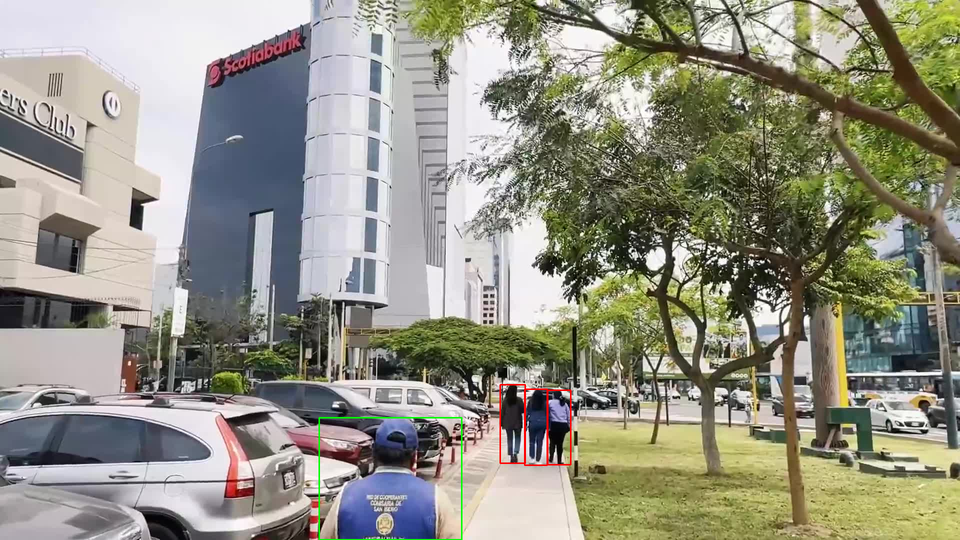}
\caption{}
\end{subfigure}
\hfill
\begin{subfigure}{0.32\linewidth}
\centering
\includegraphics[width=\linewidth]{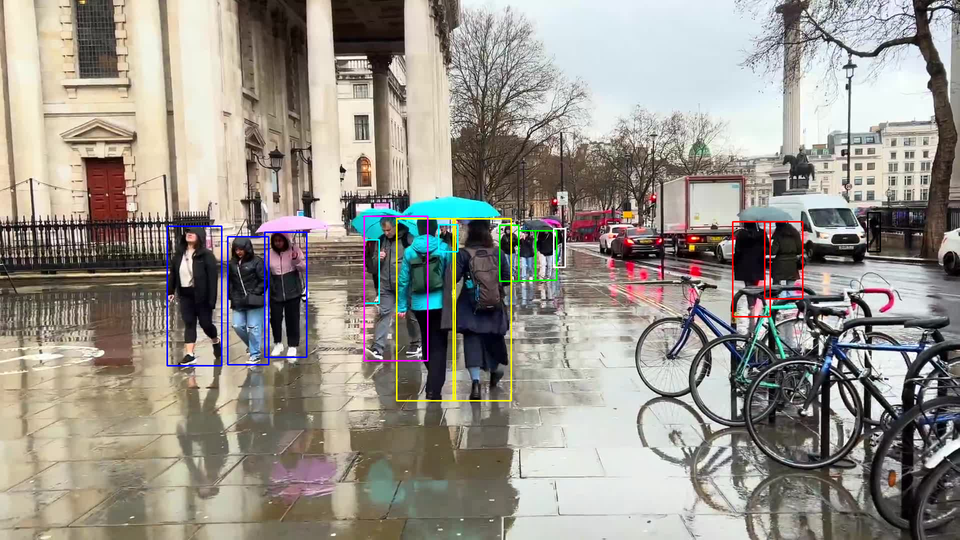}
\caption{}
\end{subfigure}
\hfill
\begin{subfigure}{0.32\linewidth}
\centering
\includegraphics[width=\linewidth]{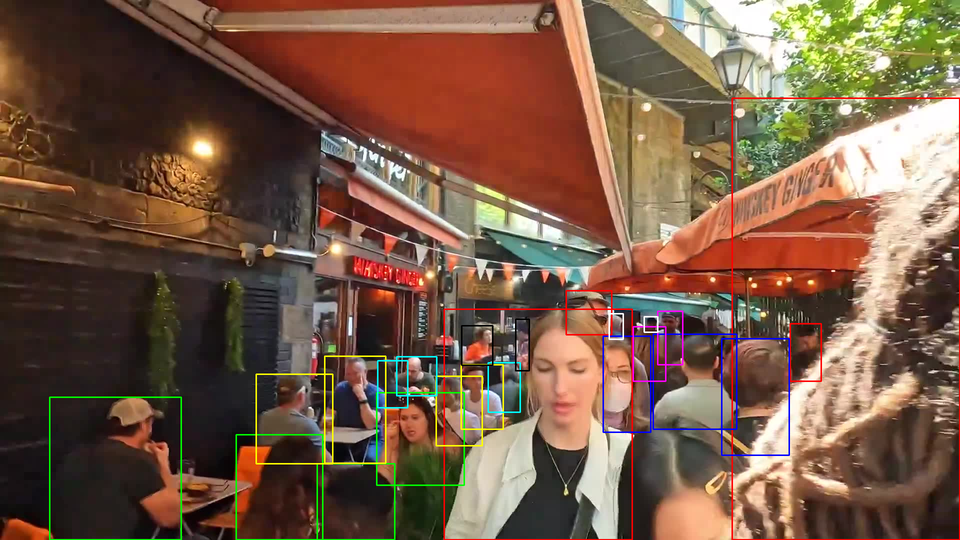}
\caption{}
\end{subfigure}

\caption{Qualitative results across frames with varying crowd densities. From left to right: (a) scattered, (b) moderate, and (c) crowded. From top to bottom: ground truth, Nvidia-Cosmos2 8B, Qwen2.5 72B, Qwen3 30B, and Gemini-3-Pro. Detections with the same color are judged to be in a social group.}

\label{fig:sup_qualitative_results}

\end{figure*}

~\autoref{fig:sup_qualitative_results} shows additional qualitative results of our approach across frames with varying crowd densities (scattered, moderate, and crowded) and different models (Nvidia-Cosmos2 8B, Qwen2.5-72B, Qwen3-30B, Gemini-3-Pro) along with the groundtruth (top) for our \dataset{}. Detections with the same color are judged to be in the same social group. We observe that model variants with fewer parameters often struggle to detect reasonable grouping in moderate and crowded scenarios.

\clearpage
\newpage

\section{Prompt Details}
\begin{figure*}[h!]
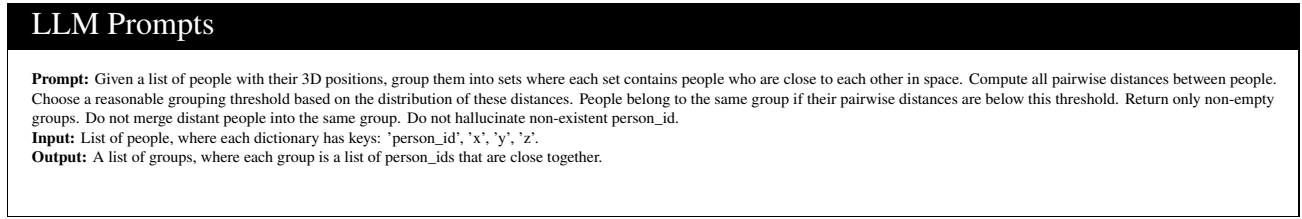

\centering

\begin{tcolorbox}[
    title={LLM Prompts},
    width=\textwidth,
    colback=white,
    colframe=black,
    boxrule=0.5pt,
    sharp corners,
    left=6pt, right=6pt, top=6pt, bottom=6pt
]

\tiny

\textbf{Prompt:} Given a list of people with their 3D positions, group them into sets where each set contains people who are close to each other in space. Compute all pairwise distances between people. Choose a reasonable grouping threshold based on the distribution of these distances. People belong to the same group if their pairwise distances are below this threshold. Return only non-empty groups. Do not merge distant people into the same group. Do not hallucinate non-existent person\_id. \\
\textbf{Input:} List of people, where each dictionary has keys: 'person\_id', 'x', 'y', 'z'. \\
\textbf{Output:} A list of groups, where each group is a list of person\_ids that are close together. \\[4pt]
\end{tcolorbox}
\vspace{-0.4cm}
\caption{Prompt instruction and input/output field descriptions processed by LLM.}
\label{fig:llm_prompt_details}
\end{figure*}

\begin{figure*}[h!]
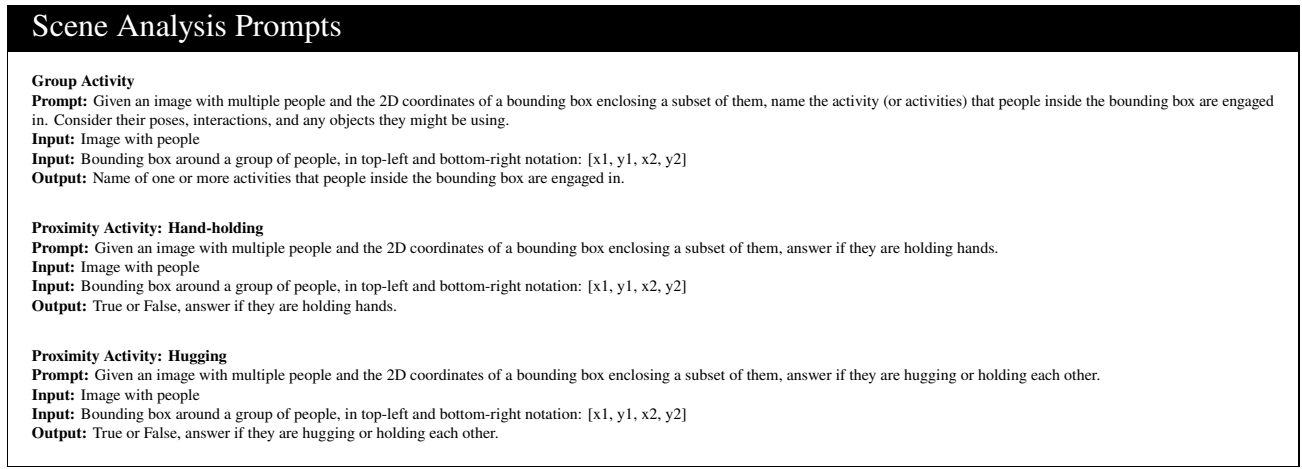

\centering
\begin{tcolorbox}[
    title={Scene Analysis Prompts},
    width=\textwidth,
    colback=white,
    colframe=black,
    boxrule=0.5pt,
    sharp corners,
    left=6pt, right=6pt, top=6pt, bottom=6pt
]
\tiny
\textbf{Group Activity} \\
\textbf{Prompt:} Given an image with multiple people and the 2D coordinates of a bounding box enclosing a subset of them, name the activity (or activities) that people inside the bounding box are engaged in. Consider their poses, interactions, and any objects they might be using. \\
\textbf{Input:} Image with people \\
\textbf{Input:} Bounding box around a group of people, in top-left and bottom-right notation: [x1, y1, x2, y2] \\
\textbf{Output:} Name of one or more activities that people inside the bounding box are engaged in. \\[4pt]

\textbf{Proximity Activity: Hand-holding} \\
\textbf{Prompt:} Given an image with multiple people and the 2D coordinates of a bounding box enclosing a subset of them, answer if they are holding hands. \\
\textbf{Input:} Image with people \\
\textbf{Input:} Bounding box around a group of people, in top-left and bottom-right notation: [x1, y1, x2, y2] \\
\textbf{Output:} True or False, answer if they are holding hands. \\[4pt]

\textbf{Proximity Activity: Hugging} \\
\textbf{Prompt:} Given an image with multiple people and the 2D coordinates of a bounding box enclosing a subset of them, answer if they are hugging or holding each other. \\
\textbf{Input:} Image with people \\
\textbf{Input:} Bounding box around a group of people, in top-left and bottom-right notation: [x1, y1, x2, y2] \\
\textbf{Output:} True or False, answer if they are hugging or holding each other.
\end{tcolorbox}
\vspace{-0.4cm}
\caption{Prompt instruction and input/output field descriptions processed by VLM for the results in Figure~\ref{fig:proximity_stats}.}
\label{fig:scene_analysis_details}
\end{figure*}

Finally, Figures \ref{fig:llm_prompt_details}, \ref{fig:vlm_prompt_details_1},  and \ref{fig:scene_analysis_details} provide detailed text descriptions of the prompts for LLMs and VLMs, following our inference code, structured using the DSPy framework. 

\begin{figure*}[h!]
\centering
\begin{tcolorbox}[
    title={VLM Prompts (1/2)},
    width=\textwidth,
    colback=white,
    colframe=black,
    boxrule=0.5pt,
    sharp corners,
    left=6pt, right=6pt, top=6pt, bottom=6pt
]
\tiny
\textbf{Prompt:} Given a list of people with their 3D positions, group them into sets where each set contains people who are close to each other in space. Compute all pairwise distances between people. Choose a reasonable grouping threshold based on the distribution of these distances. People belong to the same group if their pairwise distances are below this threshold. Return only non-empty groups. Do not merge distant people into the same group. \\
\textbf{Input:} Image with people to group \\
\textbf{Input:} List of people, where each dictionary has keys: 'person\_id', 'x', 'y', 'z'. \\
\textbf{Output:} A list of groups, where each group is a list of person\_ids that are close together. \\[4pt]

\end{tcolorbox}
\vspace{-0.4cm}
\caption{Prompt instruction and input/output field descriptions processed by VLMs.}
\label{fig:vlm_prompt_details_1}
\end{figure*}

\vspace{-0.4cm}

\end{document}